\documentclass{article}

     \usepackage[final]{neurips_2025}

\usepackage{amsmath,amsfonts,bm}

\def\eqref#1{equation~\ref{#1}}

\def\1{\bm{1}}

\DeclareMathAlphabet{\mathsfit}{\encodingdefault}{\sfdefault}{m}{sl}
\SetMathAlphabet{\mathsfit}{bold}{\encodingdefault}{\sfdefault}{bx}{n}

\newcommand{\R}{\mathbb{R}}

\newcommand{\cut}[1]{}

\usepackage[framemethod=TikZ]{mdframed}
\mdfdefinestyle{MyFrame}{%
    linecolor=black,
    outerlinewidth=0.5pt,
    roundcorner=1pt,
    innertopmargin=2pt, %
    innerbottommargin=2pt, %
    innerrightmargin=7pt,
    innerleftmargin=7pt,
    backgroundcolor=black!0!white}

\mdfdefinestyle{MyFrame2}{%
    linecolor=white,
    outerlinewidth=1pt,
    roundcorner=2pt,
    innertopmargin=10pt,
    innerbottommargin=0.5pt,
    innerrightmargin=10pt,
    innerleftmargin=10pt,
    backgroundcolor=black!3!white}

\mdfdefinestyle{MyFrameEq}{%
    linecolor=white,
    outerlinewidth=0pt,
    roundcorner=0pt,
    innertopmargin=0pt,
    innerbottommargin=0pt,
    innerrightmargin=7pt,
    innerleftmargin=7pt,
    backgroundcolor=black!3!white}

\mdfdefinestyle{FrameAdded}{%
    linecolor=black,
    outerlinewidth=0pt,
    roundcorner=0pt,
    innertopmargin=0pt,
    innerbottommargin=0pt,
    innerrightmargin=0pt,
    innerleftmargin=0pt,
    backgroundcolor=PastelGreen}

\newcommand{\AND}{\texttt{{\color{pastel_purple} \textbf{AND}}}\xspace}

\definecolor{customwhite}{HTML}{FCFBF7}
\definecolor{customturq}{HTML}{1D9D79}
\definecolor{customorange}{HTML}{D96002} 
\definecolor{custombeige}{HTML}{d6c9b1} 

\definecolor{custompurple}{HTML}{AEADF0}
\definecolor{customwhite2}{HTML}{fbf9f4}
\definecolor{customblue}{HTML}{4D9DE0}

\usepackage{silence}
\usepackage[utf8]{inputenc}
\usepackage[T1]{fontenc}
\usepackage{tabularx}
\usepackage{hyperref}
\usepackage{lipsum}
\usepackage{url}
\usepackage{placeins}
\usepackage{booktabs}
\usepackage{soul}
\usepackage{amsfonts}
\usepackage{nicefrac}
\usepackage{microtype}
\usepackage{natbib}
\usepackage{proba}
\usepackage{wrapfig}
\usepackage{caption}
\usepackage{amssymb}
\usepackage{pifont}

\usepackage{subcaption}
\usepackage{float}
\usepackage{rotating}
\usepackage{etoolbox}
\usepackage{xparse}
\usepackage{grffile}
\usepackage{amsmath}
\usepackage{xspace}
\usepackage{euscript}
\usepackage{enumitem}
\usepackage{mathtools}
\usepackage{tikz}
\usepackage{tablefootnote}
\usepackage[flushleft]{threeparttable}
\usepackage{multirow}
\usepackage[title,toc,page,header]{appendix}
\usepackage{arydshln}
\usepackage{array, booktabs}
\usepackage{comment}
\usepackage{graphicx}
\usepackage{adjustbox}
\usepackage{dsfont}
\usepackage{amsmath,amsthm}
\usepackage{balance}
\usepackage{multicol}
\usepackage{nameref}
\usepackage{pdfpages}
\usepackage{braket}
\usepackage[normalem]{ulem}
\usepackage{bbding}
\usepackage[capitalise]{cleveref}   
\usepackage{xfrac}
\usepackage[usestackEOL]{stackengine}
\usepackage{indentfirst}
\usepackage{csquotes}
\usepackage{pgf}
\usepackage{varwidth}
\usepackage{verbatimbox}
\usepackage{verbatim}
\usepackage{bm}
\usepackage{algorithmic}
\usepackage[colorinlistoftodos,prependcaption,textsize=small]{todonotes}

\usepackage{empheq}
\definecolor{myblue}{rgb}{.8, .8, 1}

\definecolor{pastelblue}{RGB}{76,113,175}
\definecolor{pastelgreen}{RGB}{144,238,144}
\definecolor{pastelred}{RGB}{196,78,82}
\definecolor{pastelgrey}{RGB}{230,230,230}
\definecolor{pastelbeige}{RGB}{243,236,221}
\definecolor{pastelpurple}{RGB}{154,139,192}
\definecolor{salmon}{RGB}{250, 128, 114}
\definecolor{darkgreen}{rgb}{0,0.6,0}
\definecolor{darkred}{rgb}{0.5,0,0}
\definecolor{verylightgreen}{HTML}{F6FFF9}
\definecolor{verylightred}{HTML}{FFF4F3}
\definecolor{verylightgray}{HTML}{F4F6F6}
\definecolor{babyblueeyes}{rgb}{0.63, 0.79, 0.95}
\definecolor{lightpink}{rgb}{1.00, 0.714, 0.757}

\usepackage{tikzpeople}
\usetikzlibrary{shapes,decorations,arrows,calc,arrows.meta,fit,positioning}

\tikzset{
    -Latex,auto,node distance =1 cm and 1 cm,semithick,
    state/.style ={ellipse, draw, minimum width = 0.7 cm},
    point/.style = {circle, draw, inner sep=0.04cm,fill,node contents={}},
    bidirected/.style={Latex-Latex,dashed},
    el/.style = {inner sep=2pt, align=left, sloped}
}

\makeatletter
\def\thmt@refnamewithcomma #1#2#3,#4,#5\@nil{%
	\@xa\def\csname\thmt@envname #1utorefname\endcsname{#3}%
	\ifcsname #2refname\endcsname
	\csname #2refname\expandafter\endcsname\expandafter{\thmt@envname}{#3}{#4}%
	\fi}
\makeatother
\Crefname{conjecture}{Conjecture}{Conjectures}
\Crefname{definition}{Definition}{Definitions}
\Crefname{observation}{Observation}{Observations}
\Crefname{assumption}{Assumption}{Assumptions}
\Crefname{axiom}{Axiom}{Axioms}
\Crefname{case}{Case}{Cases}
\Crefname{claim}{Claim}{Claims}
\Crefname{conclusion}{Conclusion}{Conclusions}
\Crefname{condition}{Condition}{Conditions}
\Crefname{criterion}{Criterion}{Criteria}
\Crefname{exercise}{Exercise}{Exercises}
\Crefname{example}{Example}{Examples}
\Crefname{notation}{Notation}{Notations}
\Crefname{problem}{Problem}{Problems}
\Crefname{property}{Property}{Properties}
\Crefname{remark}{Remark}{Remarks}
\Crefname{solution}{Solution}{Solutions}
\Crefname{summary}{Summary}{Summaries}
\Crefname{motivation}{Motivation}{Motivations}
\Crefname{query}{Query}{Queries}
\crefname{algocf}{Alg.}{Algs.}
\Crefname{algocf}{Algorithm}{Algorithms}

\newcommand*\dbar[1]{\overline{\overline{\lower0.2ex\hbox{$#1$}}}}

\DeclareFontFamily{U}{BOONDOX-calo}{\skewchar\font=45 }
\DeclareFontShape{U}{BOONDOX-calo}{m}{n}{
  <-> s*[1.05] BOONDOX-r-calo}{}
\DeclareFontShape{U}{BOONDOX-calo}{b}{n}{
  <-> s*[1.05] BOONDOX-b-calo}{}
\DeclareMathAlphabet{\mathcalb}{U}{BOONDOX-calo}{m}{n}
\SetMathAlphabet{\mathcalb}{bold}{U}{BOONDOX-calo}{b}{n}
\DeclareMathAlphabet{\mathbcalb}{U}{BOONDOX-calo}{b}{n}

\usepackage{cancel}
\renewcommand{\paragraph}[1]{{\noindent \textbf{#1.}}}

\definecolor{pastel_purple}{HTML}{756FB3}
\definecolor{pastel_green}{HTML}{1D9D79}

\colorlet{PastelPurpleLight}{pastel_purple!15!white}
\colorlet{PastelGreenLight}{pastel_green!15!white}
\sethlcolor{PastelPurpleLight}
\sethlcolor{PastelGreenLight}

\ifdefined\nohyperref\else\ifdefined\hypersetup
  \definecolor{mydarkblue}{rgb}{0,0.08,0.45}
  \hypersetup{ %
    colorlinks=true,
    linkcolor=mydarkblue,
    citecolor=mydarkblue,
    filecolor=mydarkblue,
    urlcolor=mydarkblue,
    }

  \fi
\fi

\let\originalleft\left
\let\originalright\right
\renewcommand{\left}{\mathopen{}\mathclose\bgroup\originalleft}
\renewcommand{\right}{\aftergroup\egroup\originalright}

\global\long\def\K{\mathcal{K}}

\global\long\def\inner#1#2{\left\langle #1, #2\right\rangle}

\definecolor{antiquefuchsia}{rgb}{0.57, 0.36, 0.51}
\definecolor{amethyst}{rgb}{0.6, 0.4, 0.8}

\newcommand{\deriv}[2]{\frac{\partial #1}{\partial #2}}

\newcommand{\mean}{\mathbb{E}}
\newcommand{\var}{{\rm I\kern-.3em D}}

\newcommand{\cond}{\,|\,}

\newtheorem*{theorem*}{Theorem}
\newtheorem*{proposition*}{Proposition}
\newtheorem*{example*}{Example}
\DeclareMathSymbol{\shortminus}{\mathbin}{AMSa}{"39}
\usepackage[many]{tcolorbox}

\usepackage[framemethod=TikZ]{mdframed}
\mdfdefinestyle{mybox}{%
    linecolor=white,
    outerlinewidth=1pt,
    roundcorner=2pt,
    innertopmargin=10pt,
    innerbottommargin=0.5pt,
    innerrightmargin=10pt,
    innerleftmargin=10pt,
    backgroundcolor=black!3!white}

\tcbuselibrary{theorems}

\newtcbtheorem[number within=section]{mytheo}{Statement}%
{colback=green!5,colframe=green!35!black,fonttitle=\bfseries}{th}

\newtcbox{\hlroundedpurple}[1][PastelPurpleLight]{ %
  on line,
  arc=4pt, %
  colback=#1,
  colframe=#1,
  boxrule=0pt,
  boxsep=0pt,
  left=1pt, %
  right=1pt, %
  top=2pt, %
  bottom=2pt, %
  leftrule=0pt,
  rightrule=0pt,
  toprule=0pt,
  bottomrule=0pt,
}

\newtcbox{\hlroundedgreen}[1][PastelGreenLight]{ %
  on line,
  arc=4pt, %
  colback=#1,
  colframe=#1,
  boxrule=0pt,
  boxsep=0pt,
  left=1pt, %
  right=1pt, %
  top=2pt, %
  bottom=2pt, %
  leftrule=0pt,
  rightrule=0pt,
  toprule=0pt,
  bottomrule=0pt,
}

\newtcbox{\hlrounded}[1][PastelPurpleLight]{ %
  arc=4pt, %
  colback=#1,
  colframe=#1,
  boxrule=0pt,
  boxsep=0pt,
  left=1pt, %
  right=1pt, %
  top=2pt, %
  bottom=2pt, %
  leftrule=5pt,
  rightrule=5pt,
  toprule=5pt,
  bottomrule=5pt,
}

\makeatletter
\let\save@mathaccent\mathaccent
\newcommand*\if@single[3]{%
  \setbox0\hbox{${\mathaccent"0362{#1}}^H$}%
  \setbox2\hbox{${\mathaccent"0362{\kern0pt#1}}^H$}%
  \ifdim\ht0=\ht2 #3\else #2\fi
  }
\newcommand*\rel@kern[1]{\kern#1\dimexpr\macc@kerna}
\newcommand*\widebar[1]{\@ifnextchar^{{\wide@bar{#1}{0}}}{\wide@bar{#1}{1}}}
\newcommand*\wide@bar[2]{\if@single{#1}{\wide@bar@{#1}{#2}{1}}{\wide@bar@{#1}{#2}{2}}}
\newcommand*\wide@bar@[3]{%
  \begingroup
  \def\mathaccent##1##2{%
    \let\mathaccent\save@mathaccent
    \if#32 \let\macc@nucleus\first@char \fi
    \setbox\z@\hbox{$\macc@style{\macc@nucleus}_{}$}%
    \setbox\tw@\hbox{$\macc@style{\macc@nucleus}{}_{}$}%
    \dimen@\wd\tw@
    \advance\dimen@-\wd\z@
    \divide\dimen@ 3
    \@tempdima\wd\tw@
    \advance\@tempdima-\scriptspace
    \divide\@tempdima 10
    \advance\dimen@-\@tempdima
    \ifdim\dimen@>\z@ \dimen@0pt\fi
    \rel@kern{0.6}\kern-\dimen@
    \if#31
      \overline{\rel@kern{-0.6}\kern\dimen@\macc@nucleus\rel@kern{0.4}\kern\dimen@}%
      \advance\dimen@0.4\dimexpr\macc@kerna
      \let\final@kern#2%
      \ifdim\dimen@<\z@ \let\final@kern1\fi
      \if\final@kern1 \kern-\dimen@\fi
    \else
      \overline{\rel@kern{-0.6}\kern\dimen@#1}%
    \fi
  }%
  \macc@depth\@ne
  \let\math@bgroup\@empty \let\math@egroup\macc@set@skewchar
  \mathsurround\z@ \frozen@everymath{\mathgroup\macc@group\relax}%
  \macc@set@skewchar\relax
  \let\mathaccentV\macc@nested@a
  \if#31
    \macc@nested@a\relax111{#1}%
  \else
    \def\gobble@till@marker##1\endmarker{}%
    \futurelet\first@char\gobble@till@marker#1\endmarker
    \ifcat\noexpand\first@char A\else
      \def\first@char{}%
    \fi
    \macc@nested@a\relax111{\first@char}%
  \fi
  \endgroup
}
\makeatother

\usepackage{fancyhdr}
\usepackage[utf8]{inputenc} %
\usepackage[T1]{fontenc}    %
\usepackage{hyperref}       %
\usepackage{url}            %
\usepackage{booktabs}       %
\usepackage{amsfonts}       %
\usepackage{nicefrac}       %
\usepackage{microtype}      %
\usepackage{tabularx}
\usepackage{thm-restate}
\usepackage[ruled,vlined]{algorithm2e}

\usepackage{pgfplots}
\usepackage{siunitx}

\usepackage{titletoc}

\newcommand{\dataname}{ManyPeptidesMD\xspace}
\newcommand{\name}{\textsc{Prose}\xspace}
\newcommand{\tempseq}{\texttt{RLMM}\xspace}

\newcommand{\emetric}{\(E\text{‑}\mathcal{W}_2\)\xspace}
\newcommand{\torusmetric}{\(\mathcal{T}\text{‑}\mathcal{W}_2\)\xspace}
\newcommand{\ticametric}{\(\mathrm{TICA}\text{‑}\mathcal{W}_2\)\xspace}
\newcommand{\ticajsd}{\(\mathrm{TICA}\text{‑}k\text{-}\mathrm{JSD}\)\xspace}
\newcommand{\torusjsd}{\(\mathcal{T}\text{‑}k\text{-}\mathrm{JSD}\)\xspace}
\newcommand{\sci}[2]{\(#1\cdot10^{#2}\)}
\newcommand{\scione}[1]{\(10^{#1}\)}

\title{Amortized Sampling with Transferable \\ Normalizing Flows}

\author{%
  Charlie B. Tan$^{*}$\,~$^{1}$ 
  \And
  Majdi Hassan$^{*}$\,~$^{2,3}$
  \And
  Leon Klein\,~$^{4}$
  \AND
  Saifuddin Syed\,~$^{1}$
  \And
  Dominique Beaini\,~$^{2,3,5}$
  \And
  Michael M. Bronstein\,~$^{1,6}$
  \AND
  Alexander Tong$^{\dagger}$\,~$^{2,3,6}$
  \And  
  Kirill Neklyudov$^{\dagger}$\,~$^{2,3,7}$
  \AND
  \textnormal{
  $^{1}$University of Oxford \quad
  $^{2}$Université de Montréal \quad
  $^{3}$Mila - Quebec AI Institute 
  }
  \AND
  \textnormal{
  $^{4}$Freie Universit\"at Berlin \quad
  $^{5}$Valence Labs \quad
  $^{6}$AITHYRA \quad
  $^{7}$Institut Courtois
}
}

\makeatletter
\def\blfootnote{\xdef\@thefnmark{}\@footnotetext}
\makeatother

\begin{document}

\maketitle

\begin{abstract}

Efficient equilibrium sampling of molecular conformations remains a core challenge in computational chemistry and statistical inference. Classical approaches such as molecular dynamics or Markov chain Monte Carlo inherently lack \mbox{\emph{amortization}}; the computational cost of sampling must be paid in full for each system of interest. The widespread success of generative models has inspired interest towards overcoming this limitation through learning sampling algorithms. Despite performing competitively with conventional methods when trained on a single system, learned samplers have so far demonstrated limited ability to transfer across systems. We demonstrate that deep learning enables the design of scalable and transferable samplers by introducing \name, a 285 million parameter all-atom \emph{transferable} normalizing flow trained on a corpus of peptide molecular dynamics trajectories up to 8 residues in length. \name draws zero-shot uncorrelated proposal samples for arbitrary peptide systems, achieving the previously intractable transferability across sequence length, whilst retaining the efficient likelihood evaluation of normalizing flows. Through extensive empirical evaluation we demonstrate the efficacy of \name as a proposal for a variety of sampling algorithms, finding a simple importance sampling-based fine-tuning procedure to achieve competitive performance to established methods such as sequential Monte Carlo. We open-source the \name codebase, model weights, and training dataset, to further stimulate research into amortized sampling methods and objectives.

\end{abstract}

\blfootnote{$^*$Equal contribution. $^{\dagger}$Equal advising.}
\blfootnote{Correspondence to: \texttt{charlie.tan@exeter.ox.ac.uk} and \texttt{majdi.hassan@mila.quebec}}

\section{Introduction}
\label{sec:introduction}
\begin{figure}[t]
    \vspace{-10pt}
    \centering
    \includegraphics[width=0.95\linewidth, trim={0cm 0cm 0 0}, clip]{media/visual_abstract.pdf}
    \caption{
    \label{fig:usvsmd}
    {\bf \name exceeds the quantitative performance of molecular dynamics on \emph{unseen} peptide systems.} Wasserstein-2 distances on energy, dihedral torus, and TICA projection with respect to reference molecular dynamics (\SI{5}{\us}), for a (\SI{1}{\us}) molecular dynamics baseline and \name (with SNIS), at a range of energy evaluation (above) and GPU walltime budgets (below). Each value represents the mean over 30 unseen tetrapeptide systems. \name outperforms the baseline with respect to energy evaluations on all metrics. Whilst comparable on \emetric for a given time budget, the baseline is significantly inferior on the \torusmetric and \ticametric macrostructure metrics, highlighting long simulation periods were required to traverse the separated metastable states.}
    \vspace{-10pt}
\end{figure}

Accurately sampling molecular configurations from the Boltzmann distribution is a fundamental problem in statistical physics with profound implications for understanding biological and chemical systems. Key applications include protein folding \citep{noe_constructing_2009, lindorff-larsen_how_2011}, protein–ligand binding \citep{buch_complete_2011}, and crystal structure prediction \citep{kohler_rigid_2023}; processes that underpin advances in drug discovery and material science. 

Conventional approaches such as Markov chain Monte Carlo (MCMC) \citep{liu2001monte} and, in particular, Molecular Dynamics (MD) \citep{leimkuhler2015molecular} seek to tackle this problem by proposing a general solution, which, however, has practical limitations due to its Markov nature. To accurately integrate the corresponding Hamiltonian dynamics, MD has to be simulated with a fine time-discretization (on the order of femtoseconds), which produces highly correlated samples and prevents efficient exploration of the modes of the Boltzmann density. Although running multiple chains from different initializations is possible, every chain has to be simulated for a long time to ensure proper mixing, which cannot be efficiently parallelized. Finally, the entire simulation has to be started \emph{from scratch} for a new system, which bottlenecks the speed of ab initio studies.

Deep learning-based samplers abandon the Markov chain approach to drawing samples and shift the computational burden to a one-time training phase, enabling fast and inexpensive inference compared to MCMC. In the most challenging scenario, these methods consider having access only to the unnormalized density function (analogous to MC methods) \citep{vargas2023denoising,akhound2024iterated}. Boltzmann generators (BGs) \citep{noe2019boltzmann} consider a more practical scenario where, in addition to the unnormalized density, a dataset of MD trajectory is available, which does not necessarily match the target density. To mitigate the error introduced by imperfections of the model and training data, BGs train likelihood-based models and perform self-normalized importance sampling (SNIS) \citep{liu2001monte} at inference time. The availability of training data coupled with SNIS have enabled BGs to generalize across dipeptide systems \citep{klein_transferable_2024}, but they have not yet been able to generalize across larger and more diverse systems of scientific interest.

In this work, we introduce \name, a large-scale normalizing flow which demonstrates unprecedented ability to transfer to previously unseen systems of varying amino acid composition, sequence length, and temperatures, outperforming MD for the same computational budget (see \cref{fig:usvsmd}). Our approach is strikingly simple and scalable, which elucidates the potential of the deep learning-based samplers for sampling applications. In particular, we outline the following series of contributions:
\begin{itemize}
    \item We introduce \dataname: a novel dataset of molecular dynamics trajectories for peptide systems between 2 and 8 residues. The training dataset consists of 21,700 peptide sequences simulated for \SI{200}{\ns} each, giving a total of \SI{4.3}{\ms} of simulation.
    \item Building on the recently proposed TarFlow \citep{zhai_normalizing_2024}, we propose architectural modifications, which allow for better modeling of peptide systems, system-transferable conditioning, and generation of peptide sequences of varying length.
    \item We study the use of \name as a proposal distribution for different Monte Carlo algorithms, finding the learned proposal to be sufficiently powerful for accurate sampling with standard SNIS, which does not require tuning of sampling parameters. Furthermore, resampled generations can be used for efficient fine-tuning of \name on previously unseen systems.
    \item Finally, we empirically demonstrate that \name achieves state-of-the-art performance when sampling from the equilibrium distribution on previously unseen peptide systems of length up to 8 residues surpassing the continuous normalizing flow-based transferable Boltzmann generator \citep{klein_transferable_2024} whilst generating proposals \(4\cdot10^3\) times faster.
    \item We open source our codebase \url{https://github.com/transferable-samplers/transferable-samplers}, \dataname dataset \url{https://huggingface.co/datasets/transferable-samplers/many-peptides-md} and model weights \url{https://huggingface.co/transferable-samplers/model-weights}.
\end{itemize}

\section{Background}

\subsection{Normalizing flows}
\label{sec:nf}

The fundamental challenge of probabilistic modeling is designing a density model from which samples can be efficiently generated. Normalizing flows \citep{rezende2015variational} approach this challenge by defining a diffeomorphism; a differentiable invertible function with a differentiable inverse. Namely, given a simple prior density $q_z(z)$ and a parameterized flow (diffeomorphism) $f_\theta(x)$, one can define the push-forward distribution as the map of samples from the prior distribution $z \sim q_z(z)$ via the inverse flow $x = f^{-1}_\theta(z) \sim q_\theta(x)$ with learnable parameters $\theta$. The density of the push-forward distribution can then be computed via the change-of-variables formula
\begin{align}
    q_\theta(x) ~&= \int dz\; q_z(z)\delta(x-f^{-1}_\theta(z)) = q_z(f_\theta(x))\bigg|\deriv{f_\theta(x)}{x}\bigg|\,,
    \label{eq:change_of_variables}
\end{align}
where $|\partial f_\theta(x)/ \partial x|$ is the Jacobian determinant of the map $f_\theta$. In practice, one has to be able to efficiently evaluate $f^{-1}_\theta(z)$ for sample generation, and $|\partial f_\theta(x)/ \partial x|$ for likelihood evaluation.

Autoregressive normalizing flows \citep{kingma2016improved,papamakarios2017masked,zhai_normalizing_2024} define a family of invertible maps with tractable Jacobian as a sequence of composed transformations $f_\theta = f_\tau \circ \ldots \circ f_0$, where each transformation $z_{t+1} = f_t(z_{t})$, $z_0=x$ is defined autoregressively. In the case of the TarFlow \citep{zhai_normalizing_2024}, this is an autoregressive affine update defined over blocks of latent variable corresponding to image patches $z_t[i] \in \R^D$. That is, the $i$-th latent block is
\begin{align}
z_{t+1}[i] =
    \begin{cases}
    z_t[i]\,, & i=0\,,\\
    (z_t[i] - \mu_t(z_t[:i])[i]) \odot \exp(- \alpha_t(z_t[:i])[i]) \,, & i \in [1,N-1]\,,
    \end{cases}
    \label{eq:tarflow_fwd}
\end{align}
where we adopt slicing notation denoting the $i$-th block as $z[i]$ and blocks up to the $i$-th (exclusive) as $z[:i]$. Notably, the autoregressive structure allows for efficient evaluation of the Jacobian determinant due to its lower-triangular structure $\log|\partial f_t(z_{t})/\partial{z_{t}}| = - \sum_{i=1}^{N-1} \sum_{j=0}^{D-1} \alpha_t(z_{t}[:i])[i]_j$. Furthermore, the autoregressive affine updates are invertible with inverse $z_{t} = f^{-1}_{t}(z_{t+1})$ given by
\begin{align}
    z_{t}[i] =
    \begin{cases}
    z_{t+1}[i]\,, & i=0\,,\\
    z_{t+1}[i] \odot \exp(\alpha_{t}(z_{t}[:i])[i]) + \mu_{t}(z_{t}[:i])[i]\,, & i \in [1,N-1]\,.
    \end{cases}
    \label{eq:tarflow_rev}
\end{align}

However, clearly, such transformations leave the leading dimension $z_{t}[0]$ untouched, hence must be interleaved with permutations \(\pi_t\) over the dimensions $f_\theta = \pi_{\tau}\circ f_\tau\circ \ldots \circ \pi_{0} \circ f_0$. For example, \citet{zhai_normalizing_2024} use simple inversions of the latent block sequence for all $\pi_t$ across the entire model.

\subsection{Boltzmann generators}

Despite normalizing flows allowing various forms of training supervision, such as variational inference for unnormalized densities \citep{rezende2015variational}, or maximum likelihood from empirical distributions \citep{kingma2018glow}, errors present in the parameterized distribution prevents accurate evaluation within scientific applications requiring high precision, e.g. free energy estimation. \textit{Boltzmann generators} address specifically this challenge by performing self-normalized importance sampling (SNIS) at inference time. Namely, to evaluate the expectation of a statistic $\varphi(x)$ w.r.t. the target Boltzmann density $p(x)$ one can use the following consistent Monte Carlo estimator 
\begin{align}
    \mean_{p(x)}\varphi(x) \approx \sum_{i=1}^n \frac{w_i}{\sum_{j=1}^n w_j} \varphi(x_i)\,,\;\; w_i = \frac{p(x_i)}{q_\theta(x_i)}\,,\;\; x_i \sim q_\theta(x)\,,
\end{align}
where $q_\theta(x)$ is the density of the learned normalizing flow. The SNIS estimator converges, for $n \to \infty$, to the true value $\mean_{p(x)}\varphi(x)$. Furthermore, \citet{tan_scalable_2025} extended the Boltzmann generator framework to more general Monte Carlo algorithms, in particular a continuous-time formulation of Sequential Monte Carlo \citep{jarzynski1997nonequilibrium, albergo2024nets}.

Transferable Boltzmann generator (TBG) \citep{klein_transferable_2024} made a first attempt of learning a sampler that generalizes across target densities corresponding to different peptide systems. TBG parametrizes the proposal distribution as a continuous normalizing flow (CNF) \citep{chen2018neural}, where the vector field is defined by an equivariant graph neural network \citep{satorras2021n,klein_equivariant_2023}. Crucial to the method is the system-dependent conditioning of $N$ atoms
\begin{align}
    h[i] = [A_i, R_i, P_i]\,,
\end{align}
where atom type $A_i$, residue type $R_i$, and residue position $P_i$ are each encoded as one-hot vectors. Training on a set of MD trajectories for dipeptide systems \citep{klein2023timewarp} with this system-conditional encoding enables TBG to generate a proposal for previously unseen dipeptides.

However, despite successful generalization across dipeptides, the TBG architecture introduces significant bottlenecks for inference and fine-tuning. Indeed, the learned CNF requires accurate integration of the vector field and computationally expensive evaluation of its divergence for evaluating the learned density model. For instance, the implementation of \citep{klein_transferable_2024} requires 4 GPU-days to produce $3\times10^4$ samples with their corresponding proposal likelihoods for a single dipeptide system. Furthermore, the expensive evaluation of density makes it infeasible to train or finetune TBG via the reverse KL-divergence or create a replay buffer of a substantial size.

\section{Scalable transferable normalizing flows as Boltzmann generators}\label{sec:methods}

\subsection{Architecture of \name}
\label{sec:nf_architecture}

\name builds on the TarFlow architecture \citep{zhai_normalizing_2024}, which parametrizes a sequence of autoregressive affine transformations via blocks of transformer layers. The expressivity and favorable scalability of the transformer layers enables TarFlow to effectively model high dimensional data, whilst the affine autoregressive flow parameterization ensure fast and accurate energy evaluation. With minimal modifications TarFlow is capable of successfully modeling high-dimensional molecular data \citep{tan_scalable_2025}. Here we describe our design choices that make transferability possible.

\paragraph{Transferability across system dimensions} We extend TarFlow to support concurrent training on sequences of variable length. Whilst transformers natively support sequences of arbitrary length, special consideration is required within a normalizing flow such as TarFlow that is defined for fixed input and output dimensions. We therefore define appropriate masking to the affine sequence updates and log-determinant aggregation to prevent padding tokens influencing either computation, under arbitrary sequence permutations; further details of the tokenization and masking are provided in \cref{app:architecture}. We additionally replace the fixed-length learnable position embedding with the more extrapolation-friendly sinusoidal embedding. This design enables \name to efficiently train across a distribution of systems $s$ by maximizing the normalized log-likelihood
\begin{align}
    \max_\theta \mean_{s} \frac{1}{d(s)}\mean_{x \sim p(x\cond s)} \log q_\theta(x)\,
\end{align}
where $d(s)$ is the size of the system $s$ \citep{klein_transferable_2024}. This extended architecture allows for parallel processing of data dimensions, enabling transferability and scalability across lengths.

\paragraph{Adaptive system conditioning} The standard TarFlow employs simple additive conditioning for class-conditional image generation. Whilst we find this to be sufficient to define a system-transferable normalizing flow, we follow large-scale atomistic transformer architectures in applying conditioning through adaptive layer normalization, adaptive scaling, and SwiGLU transition blocks~\citep{abramson_accurate_2024, geffner_proteina_2024}. The system conditioning features are constructed from atom types \(A\), residue types \(R\), residue positions \(P\), and sequence lengths \(L\). Atom and residue types are embedded using lookup-table embedding layers, whilst sinusoidal embeddings are employed for the naturally ordered sequence position and sequence length. See \cref{app:architecture} for further details.

\paragraph{Chemistry-aware sequence permutations} In the image setting, \citet{zhai_normalizing_2024} employ only an identity and flip permutation to the sequence of image tokens. Similarly, when applying TarFlow to peptide systems \citet{tan_scalable_2025} employ only an identity and flip permutation on the ordering defined per-residue starting with backbone atoms followed by sidechain atoms. Whilst a simple identity and flip  may be appropriate for the regular grid of image data, we argue this to be suboptimal for the diversity of geometric interactions present in molecular systems. This motivates our introduction of chemistry-aware sequence permutations, defined to promote effective peptide modeling. We define the \emph{backbone permutation}, such that the backbone atoms \([N_i,C_{\alpha,i},C_i,O_i]_{i=0}^{L-1}\) (with associated hydrogens) for all residues are located at the start of the sequence, and followed by the sidechains. By processing the coordinates of the backbone atoms at the start of the sequence, the model refines the global structure of the peptide as a contiguous sequence. Crucially, the sidechain positions are subsequently able to causally attend to the full backbone structure, hence enabling local updates to be influenced by global structure. We further employ a \emph{backbone-flip} permutation to provide additional diversity to the autoregressive modeling.

\begin{figure}
\vspace{-10pt}
    \centering
    \includegraphics[width=\linewidth, trim=5cm 6cm 0cm 6cm, clip]{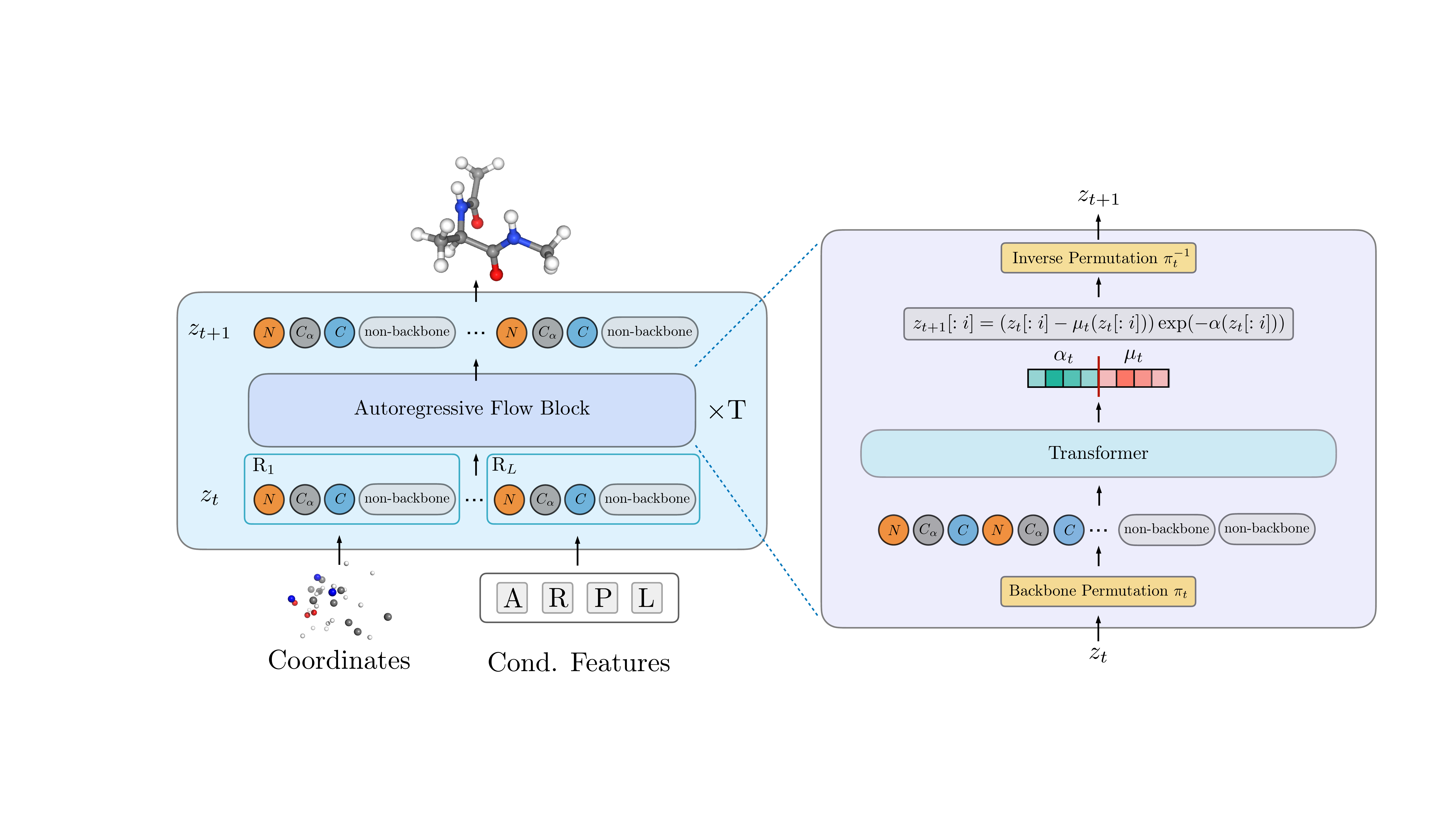}
    \caption{{\bf All-atom block-wise autoregressive normalizing flow based on the TarFlow \citep{zhai_normalizing_2024}.} Peptides are encoded via atom types \(A\), residue types \(R\), residue position \(P\), and sequence length \(L\). Atom positions in 3D Cartesian coordinates define the system state. The embedding of the peptide is applied as conditioning to the coordinates such that \name achieves transferability between systems. Within each block the sequence \(z_t\) is permuted and passed to a transformer, defining an autoregressive affine update. In the backbone permutation the backbone \([N_i,C_{\alpha,i},C_i, O_i]_{i=0}^{L-1}\) of all residues (with associated hydrogens) is updated before any sidechains, providing additional diversity to the causal attention for global structure modeling.}
    \label{fig:enter-label}
\vspace{-10pt}
\end{figure}

\subsection{Inference and fine-tuning of \name}
\label{sec:inference_methods}
The practical applicability of learned sampling methods depends significantly on their inference-time throughput, as well as their transferability to unseen systems. In this section, we describe how one can employ \name for inference-time importance sampling, importance sampling-based fine-tuning, as well as annealing of the learned proposal to different target temperatures.

\textbf{Importance sampling.} At the inference time, one can use \name to estimate the expectation of statistics $\varphi(x)$ w.r.t. the target Boltzmann density $p(x)$ via a self-normalized importance sampling (SNIS) estimator. Namely, we consider standard SNIS, discrete-time sequential Monte Carlo (SMC) \citep{neal2001annealed,doucet2001sequential}, and continuous-time SMC \citep{jarzynski1997nonequilibrium,albergo2024nets}. All these estimators are of the form
\begin{align}
    \mean_{p(x)}\varphi(x) \approx \sum_{i=1}^n \frac{w_i}{\sum_{j=1}^n w_j}\varphi(x_i)\,,\;\; w_i = \frac{p(x_i)}{q(x_i)}\,,\;\; x_i \sim q(x)\,,
    \label{eq:snis}
\end{align}
where the only difference between them is the proposal density $q(x)$.
Note that these estimators can be interpreted as the expectation over the empirical distribution, i.e.
\begin{align}
    \sum_{i=1}^n \frac{w_i}{\sum_{j=1}^n w_j}\varphi(x_i) = \mean_{\tilde{p}(x)}\varphi(x)\,,\;\; \tilde{p}(x) = \sum_{i=1}^n \frac{w_i}{\sum_{j=1}^n w_j}\delta(x-x_i)\,.
\end{align}
In practice, we compare the true density $p(x)$ with our generated distribution $\tilde{p}(x)$ instead of measuring statistics $\varphi(x)$. For completeness, we describe all the considered estimators in \cref{app:importance_sampling}.

\textbf{Self-improvement.} For an unseen system $s$ we demonstrate the ability to fine-tune \name using a self-improvement strategy. Namely, we iteratively generate the empirical distribution $\tilde{p}(x\cond s)$ by resampling the samples from the model $q_\theta(x\cond s)$ proportionally to $p(x\cond s)$ and use these samples for fine-tuning. Note that this is different from classical fine-tuning as true samples from the target are not available. We update the parameters by maximizing the likelihood on the resampled proposal, i.e.
\begin{align}
    \max_\theta \mean_{\tilde{p}(x\cond s)}\log q_\theta(x\cond s)\,,\;\; \tilde{p}(x\cond s) = \sum_{i=1}^n \frac{w_i}{\sum_{j=1}^n w_j}\delta(x-x_i)\,, w_i = \texttt{detach}\left(\frac{p(x_i \cond s)}{q_\theta(x_i\cond s)}\right)\,.
\end{align}
This is akin to the energy-based training of \citet{jing2022torsional}, in which samples are proposed by ODE-integration of a diffusion model, resampled, and then used in the score-matching objective.

\textbf{Temperature transfer.} Temperature is fundamental to molecular simulation, with significant influence on conformational dynamics and statistic expectations. It is therefore highly desirable that a learned sampler may transfer across temperature without retraining. Formally, we aim to change the temperature $T = 1/\beta$ of the learned density model when generating samples, i.e.
\begin{align}
    \beta\log q_\theta(f^{-1}_\theta(z)) ~& = \beta\log q_z(z) - \beta\log\bigg|\deriv{f^{-1}_\theta(z)}{z}\bigg|\,.
\end{align}
Note that, for measure-preserving flows $\log|\partial f^{-1}_\theta(z)/ \partial z| = 0$, one simply has to change the temperature of the prior distribution (i.e. sample $z \sim q_z(z)^\beta$ instead of $z \sim q_z(z)$) to change the temperature of the density model, which is a standard technique in the normalizing flow literature \citep{kingma2018glow, dibak2021temperature}. Whilst \name is a non-volume preserving flow \citep{dinh2016density}, hence violating this assumption, we found that simply scaling the prior temperature $\beta\log q_z(z)$ results in a suitable proposal for the Boltzmann density with the corresponding temperature.

\section{Experiments}\label{sec:experiments}

To establish the performance of \name, we first introduce a new dataset of peptide molecular dynamics. We employ this dataset to train \name and prior methods, and evaluate using metrics computed against reference molecular dynamics trajectories. We additionally evaluate \name as a proposal for a variety of sampling algorithms, and in the temperature-transfer setting.

\subsection{Molecular dynamics trajectory dataset}
\label{sec:dataset}

We introduce ManyPeptidesMD; a novel dataset of peptide MD trajectories for sequences ranging from 2 to 8 residues in length\footnote{Available at \url{https://huggingface.co/datasets/transferable-samplers/many-peptides-md}}.
Following \citet{klein2023timewarp} all simulation is performed using \texttt{OpenMM} \citep{eastman_openmm_2017} with the Amber14 forcefield \citep{amber14}. 
For training, a total of 21,700 uniformly sampled sequences are simulated for \SI{200}{\ns}.
For evaluation, 30 sequences of length 2, 4, and 8 are randomly sampled such that all amino acids are represented equally, and simulated for \SI{5}{\us}. 
Further details on dataset collection and MD configuration provided in \cref{app:dataset}. 

\begin{table}[ht!]
\vspace{-8pt}
\caption{\small Number of sequences used per peptide length for training and evaluation.}
\label{tab:length_distribution}
\centering
\begin{tabular}{@{}lccccccc@{}}
    \toprule
    Sequence length & 2 & 3 & 4 & 5 & 6 & 7 & 8 \\
    \midrule
    Training   & 200  & 1,000  & 1,500 & 2,000 & 3,000 & 4,000 & 10,000 \\
    Evaluation & 30   & ---  & 30   & ---  & ---  & ---  & 30   \\
    \bottomrule
\end{tabular}
\end{table}

\subsection{Experimental configuration}
\label{sec:experimental_config}

We train the first Boltzmann generators transferable across peptide sequence length. We train the \name architecture defined in \cref{sec:nf_architecture}, an unmodified TarFlow \citep{zhai_normalizing_2024} as in SBG~\mbox{\citep{tan_scalable_2025}}, and the equivariant CNF of \citet{klein_transferable_2024}, with the improved training recipe of \citet{tan_scalable_2025}, denoted as ECNF++. All models are trained for \(5 \times 10^5\) iterations with batch size 512. Both \name and TarFlow are suitably scalable to long sequences and are trained on the full dataset detailed in \cref{sec:dataset}. However, generating 8 residue sequences with likelihoods for ECNF++ was found to be prohibitively expensive, hence the training data was limited to sequences up to and including length 4. Comprehensive training details are provided in \cref{app:training_details}.

The primary evaluation metrics are the Wasserstein-2 distance on: (i) the energy distribution \emetric, (ii) the dihedral angle torus distribution \torusmetric, (iii) the first 2 TICA component projections \ticametric. The energy distribution is highly sensitive to perturbation in bond length and angle, hence \emetric measures accuracy on fine-grained details. The dihedral angle tori and TICA projection describe macrostructure, hence \torusmetric and \ticametric measure accuracy in terms of metastable state coverage. We additionally report effective sample size (ESS); the variance of the importance weights. For metric definitions and further details on sampling evaluation procedure please refer to \cref{app:evaluation_details}. 

\subsection{Scale transferability of \name}

To establish the performance of \name as a sampler proposal distribution, we first evaluate the trained flows in the Boltzmann generator setting. Here we generate a set of proposal particles \(\{x_i\}_{i=1}^N\), evaluate model likelihoods \(q_\theta(x_i)\) and reweight using SNIS as in \cref{eq:snis}. In addition to the trained models we benchmark against the following pretrained baselines; (i) the TBG model trained by \citet{klein_transferable_2024}, denoted as ECNF (ii) TimeWarp~\citep{klein_timewarp_2023} (iii) BioEmu~\citep{lewis2024scalable} (iv) Unisim~\citep{yu2025unisimunifiedsimulatortimecoarsened}. For all methods we permit a budget of \(10^4\) energy evaluations. For the Boltzmann generator methods (ECNF, ECNF++, TarFlow, \name) this corresponds to \(10^4\) SNIS particles; for further information on the budget allocation of non-BG methods (TimeWarp, BioEmu, UniSim) see \cref{app:evaluation_details}. We additional provide results for the unweighted proposal distributions in \cref{app:additional_results}. We note the TimeWarp dataset to lack any sequences containing Proline at the N-Terminal, hence neither TimeWarp nor ECNF were evaluated on such sequences.

\begin{table*}[ht!]
\caption{\small Quantitative results for baseline methods, and flows with self-normalized importance sampling on peptide systems up to 8 residues. All methods evaluated a budget of \(10^4\) energy evaluations. Best values in \textbf{bold}. \textsuperscript{*}Not evaluated on sequences with N-terminal proline due to absence in training data.}
\label{tab:main_results}
\resizebox{1\linewidth}{!}{
\begin{tabular}{@{}lccccccccccc}
    \toprule
    Sequence length $\rightarrow$ & \multicolumn3c{2AA \tiny{(30 systems)}} & \multicolumn4c{4AA \tiny{(30 systems)}} & \multicolumn4c{8AA \tiny{(30 systems)}}  \\
    \cmidrule(lr){2-4}\cmidrule(lr){5-8}\cmidrule(lr){9-12}
    Model $\downarrow$ & ESS $\uparrow$ & \emetric  $\downarrow$  & \torusmetric  $\downarrow$  & ESS $\uparrow$ & \emetric  $\downarrow$  & \torusmetric  $\downarrow$  & \ticametric  $\downarrow$  & ESS $\uparrow$ & \emetric  $\downarrow$  & \torusmetric  $\downarrow$  & \ticametric  $\downarrow$  \\
    \midrule
    TimeWarp\textsuperscript{*} & --- & 4.532 & 0.842 & --- & 7.237 & 2.204 & 0.993 & --- & --- & --- & --- \\
    BioEmu & --- & 45.313 & 1.208 & --- & 90.079 & 2.037 & 1.479 & --- & 193.873 & 4.638 & 1.601 \\
    UniSim & --- & $>10^{5}$ & 1.289 & --- & $>10^{4}$ & 2.766 & 1.733 & --- & $>10^{3}$ & 6.156 & 1.495 \\
    \midrule
    ECNF\textsuperscript{*} & 0.086 & 0.894 & 0.488 & --- & --- & --- & --- & --- & --- & --- & --- \\
    ECNF++ & 0.024 & 3.470 & 0.302 & 0.008 & 10.032 & 1.121 & 0.572 & --- & --- & --- & --- \\
    TarFlow & 0.134 & 0.452 & \textbf{0.193} & 0.045 & 1.260 & 0.924 & 0.492 & 0.008 & 11.298 & 2.733 & 1.087 \\
    \name & \textbf{0.191} & \textbf{0.371} & 0.210 & \textbf{0.071} & \textbf{0.932} & \textbf{0.752} & \textbf{0.367} & \textbf{0.011} & \textbf{10.038} & \textbf{2.456} & \textbf{0.988} \\
    \bottomrule
    \end{tabular}
    }
\end{table*}

\begin{figure}[t]
    \centering
    \includegraphics[width=\textwidth]{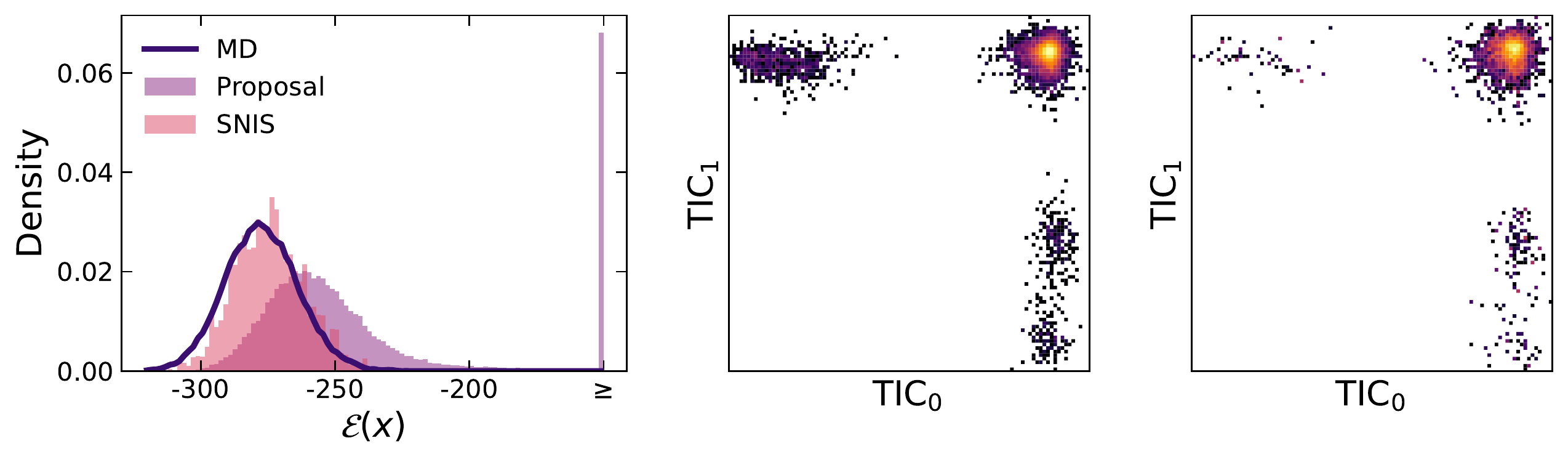}
    \caption{{\bf \name accurately samples from the Boltzmann distributions of unseen octapepitde system.} Empirical results for sampling from \texttt{DGVAHALS} peptide system, not present in training data. Energy histogram (left) for reference MD data, \name proposal and \name reweighted using SNIS, demonstrate fine-grained detail accuracy. TICA plots for MD (center) and SNIS-reweighted \name (right) illustrate mode coverage.}
    \label{fig:octopeptide}
    \vspace{-5pt}
\end{figure}

We present metrics for \name and baseline methods in \cref{tab:main_results}. \name achieves the strongest performance on all metrics aside from dipeptide \torusmetric, where it is marginally outperformed by TarFlow, confirming it to be a strong SNIS proposal for peptide systems of varying sequence length. ECNF++ performs very poorly on \emetric on both dipeptides and tetrapeptides, seemingly unable to learn an effective vector field when trained on tetrapeptides. TimeWarp is the strongest non-Boltzmann generator baseline, with both BioEmu and UniSim attaining high values of \emetric. However, we note the training data for these pretrained models does not correspond exactly with our evaluation data and hence they are not directly comparable. \cref{fig:usvsmd} further confirms the success of \name with SNIS as an amortized sampler, surpassing the performance of a baseline MD trajectory on the critical \torusmetric and \ticametric describing metastable state coverage w.r.t. both energy evaluations and GPU walltime. We additionally present qualitative results on the unseen octapeptide \texttt{DGVAHALS} in \cref{fig:octopeptide}, demonstrating the unprecedented scalability of \name; further results are provided in \cref{app:additional_results}.

\subsection{Architecture ablation study}

We proceed to ablate the architectural variations applied in \name, as described in \cref{sec:nf_architecture}; (i) the adaptive system conditioning blocks in the transformer layers (ii) the backbone permutations interleaved into our permutation sequence. We train ablation models using an identical training configuration to that described in \cref{sec:experimental_config}. We present metrics for these modifications in \cref{tab:main_ablation_results}. We observe a significant improvement in all metrics across scales of peptide sequence length, confirming the efficacy of these modifications for atomistic modeling, notably the backbone permutations which introduce negligible runtime complexity over the standard TarFlow architecture.

\begin{table*}[ht!]
\caption{\small Ablation results for \name architecture components. SNIS performed with \(2 \times 10^5\) energy evaluations.}
\label{tab:main_ablation_results}
\resizebox{1\linewidth}{!}{
\begin{tabular}{@{}lccccccccccc}
    \toprule
    Sequence length $\rightarrow$ & \multicolumn3c{2AA \tiny{(30 systems)}} & \multicolumn4c{4AA \tiny{(30 systems)}} & \multicolumn4c{8AA \tiny{(30 systems)}}  \\
    \cmidrule(lr){2-4}\cmidrule(lr){5-8}\cmidrule(lr){9-12}
    Model $\downarrow$ & ESS $\uparrow$ & \emetric  $\downarrow$  & \torusmetric  $\downarrow$  & ESS $\uparrow$ & \emetric  $\downarrow$  & \torusmetric  $\downarrow$  & \ticametric  $\downarrow$  & ESS $\uparrow$ & \emetric  $\downarrow$  & \torusmetric  $\downarrow$  & \ticametric  $\downarrow$  \\
    \midrule
    \name & \textbf{0.191} & \textbf{0.282} & 0.177 & \textbf{0.071} & \textbf{0.646} & \textbf{0.607} & \textbf{0.349} & \textbf{0.011} & \textbf{9.360} & \textbf{2.019} & \textbf{0.960} \\
    \midrule
    w/o Backbone-first & 0.170 & 0.295 & \textbf{0.152} & 0.051 & 0.816 & 0.697 & 0.421 & 0.009 & 10.261 & 2.275 & 1.044 \\
    w/o Transition & 0.152 & 0.322 & 0.282 & 0.054 & 0.880 & 0.691 & 0.384 & 0.009 & 11.384 & 2.209 & 1.012 \\
        \bottomrule
        \end{tabular}
    }
    \vspace{-5pt}
\end{table*}

\subsection{Sampling algorithms}

Having established the unmatched performance of \name in the standard Boltzmann generator framework, we now consider alternative sampling algorithms made tractable by its efficient likelihood. We evaluate SNIS, SMC in continuous time, SMC in discrete time, and the simple instantiation of self-improvement defined in \cref{sec:inference_methods}. All methods are permitted a budget of \(10^6\) energy evaluations, further details on method configurations are provided in \cref{app:evaluation_details}. Metric results are presented in \cref{tab:sampling_alg_results}. These results reveal the surprising result that, given a suitably strong proposal distribution, SNIS is competitive with both SMC variants, despite requiring no tuning. While SMC discrete achieves the best value of \emetric on octapeptides, both SMC variants introduce a notable deterioration of macrostructure metrics at this scale when compared to SNIS. Furthermore, the performance of SNIS with self-improvement at improving the tetra- and octapeptide \emetric provides strong evidence in favor of proposal fine-tuning within sampling methods, as an alternative to resource allocation solely on annealing-based methods. We provide results for the unseen \tempseq system in \cref{fig:sael_tica}, illustrating the superior mode coverage of \name with SNIS over a MD baseline given an allocation of \(10^6\) energy evaluations, further evidence of successful amortized sampling with \name.

\begin{table*}[ht!]
\caption{\small Results for samplers using \name as proposal. Methods provided with budget of \(10^6\) energy evaluations. Best values \textbf{bolded}. \textsuperscript{*}Not evaluated on sequences with N-terminal proline due to absence in training data.}
\label{tab:sampling_alg_results}
\resizebox{1\linewidth}{!}{
\begin{tabular}{@{}lcccccccccccccc}
    \toprule
    Sequence length $\rightarrow$ & \multicolumn3c{2AA \tiny{(30 systems)}} & \multicolumn4c{4AA \tiny{(30 systems)}} & \multicolumn4c{8AA \tiny{(30 systems)}}\\
    \cmidrule(lr){2-4}\cmidrule(lr){5-9}\cmidrule(lr){10-13}
    Algorithm $\downarrow$ & ESS $\uparrow$ & \emetric  $\downarrow$  & \torusmetric  $\downarrow$  & ESS $\uparrow$ & \emetric  $\downarrow$  & \torusmetric  $\downarrow$  & \ticametric  $\downarrow$ & ESS $\uparrow$ & \emetric  $\downarrow$  & \torusmetric  $\downarrow$  & \ticametric  $\downarrow$  \\
    \midrule
    Timewarp\textsuperscript{*} & --- & 2.551 & 0.580 & --- & 4.125 & 1.600 & 0.813 & --- & --- & --- & --- \\
    \midrule
    SNIS & 0.190 & 0.271 & 0.165 & 0.070 & 0.665 & 0.613 & 0.349 & 0.011 & 9.386 & \textbf{2.012} & \textbf{0.964} \\
    SMC Continuous & --- & 0.318 & 0.177 & --- & 0.764 & 0.688 & \textbf{0.338} & --- & 10.563 & 2.642 & 1.049 \\
    SMC Discrete & --- & \textbf{0.249} & \textbf{0.147} & --- & 0.653 & 0.721 & 0.400 & --- & \textbf{7.672} & 2.524 & 1.086 \\
    {Self-Improve + SNIS} & 0.189 & 0.265 & 0.171 & 0.070 & \textbf{0.568} & \textbf{0.611} & 0.345 &0.011 & 8.886 & 2.071 & 0.966\\
    \bottomrule
    \end{tabular}
}
\vspace{-5pt}
\end{table*}

\begin{figure}[t]
    \centering
    \includegraphics[width=\linewidth, trim=0cm 0cm 0cm 0.cm, clip]{media/RLMM_ticas.pdf}
    \caption{{\bf By drawing \emph{uncorrelated} proposal samples, \name achieves greater metastable state coverage than molecular dynamics for the same number of energy evaluations.} TICA projection plots for unseen tetrapeptide system (\tempseq). After \(5 \cdot 10^9\) energy evaluations the reference molecular dynamics (left) has traversed four distinct metastable states, taken to be ground truth. However, with an energy evaluation budget of \(10^6\) molecular dynamics explores only a single metastable state (center), highlighting the limitations of simulation-based sampling methods for mode exploration. \name with SNIS (right) samples all 4 states given the same budget of energy evaluations, indicating successful amortization of the mode exploration problem.}
    \label{fig:sael_tica}
    \vspace{-5pt}
\end{figure}

\subsection{Inference-time temperature transfer}

We lastly evaluate the scaled prior (SP) technique for inference-time temperature transfer introduced in \cref{sec:inference_methods}. We collect additional \SI{1}{\us} MD trajectories for the \tempseq unseen tetrapeptide at temperatures defined by geometric series between the base model temperature of \SI{310}{\K}, and \SI{800}{\K}. We then perform SNIS using \(2\times 10^5\) energy evaluations from \name, both naively and with the scaled prior inference method. Results are presented in \cref{fig:temp-result}, with scaled prior universally outperforming naive SNIS. We emphasize that scaled prior \emph{does not} require any fine-tuning and introduces negligible increase in complexity at inference. These results thus demonstrate that \name is transferable not only in system, but also in temperature, opening a variety of avenues of further exploration.

\begin{figure}[H]
    \centering
    \includegraphics[width=\linewidth]{media/temp_metrics.pdf}
    \caption{{\bf Scaled prior greatly improves the ability of \name to accurately reweight to arbitrary temperatures.}
    Metrics for \name on \tempseq unseen tetrapeptide, targeting temperatures up to \SI{800}{\K}.
    Naively applying SNIS to the target temperature leads to a rapid degradation in energy distribution, and to a lesser extent the dihedral angle distribution. Applying prior scaling (\name SP) leads to a significant improvement in energy distribution at high temperatures and moderate improvement in dihedral angles. Notably, the TICA distribution \emph{improves} at higher temperatures irrespective of scaled prior usage, although scaled prior remains more effective.}
    \label{fig:temp-result}
\end{figure}

\section{Related Work}

\paragraph{Normalizing flows} Normalizing flows~\citep{rezende2015variational, dinh2016density,kingma2018glow,durkan2019neural} fell from favor as general-purpose generative models as generative adversarial networks (GANs) \citep{goodfellow2014generativeadversarialnetworks}, diffusion models~\citep{ho2020denoisingdiffusionprobabilisticmodels,song2021scorebasedgenerativemodelingstochastic}, and continuous normalizing flows \citep{chen_neural_2018, liu_rectified_2022, albergo_building_2023}, demonstrated superior empirical generative quality. However, they have still found relevance in scientific applications where efficient likelihood calculations are necessary. Furthermore, the recent introduction of Transformer-based normalizing flows \citep{zhai_normalizing_2024, kolesnikov2024jet} has enabled previously intractable data distributions to be modeled whilst retaining efficient likelihood evaluation, and brought renewed research attention to this area.

\paragraph{Boltzmann generators} Boltzmann generators are machine learning-based samplers that train likelihood-based models and employ inference-time SNIS to achieve consistent sampling of the target density \citep{noe2019boltzmann}. A major limitation of standard Boltzmann generators is the need to train the proposal model on a dataset of true density samples, motivating methods that transfer between molecular systems. Accordingly, \citet{jing2022torsional} propose a transferable Boltzmann generator operating on the torsion angles of small molecules, and \citet{klein_transferable_2024} develop a Boltzmann generator operating on Cartesian coordinates that transfers between dipeptide systems. However, the scalability of such methods has remained limited due to the difficulty of designing expressive generative models that possess efficient and accurate likelihood evaluations. In particular the use of continuous normalizing flows implies a large cost to proposal likelihoods due to the need to integrate the vector field divergence~\citep{grathwohl_ffjord_2019}. \citet{schopmans_temperature-annealed_2025} replace the single-step SNIS of standard Boltzmann generators with a temperature-annealing sequence of normalizing flows, performing SNIS with each flow to sample from a given target distribution. 

\paragraph{Approaches to machine learning-based sampling} The widespread empirical success of generative modeling has inspired many approaches to machine learning-based sampling. Boltzmann emulators, like Boltzmann generators, seek uncorrelated sampling of the target density, but forgo efficient likelihood evaluation in favor of scalable generative modeling on large pre-collected datasets \citep{abdin_direct_2024a,waymentSteele2024,Lewis2024.12.05.626885}. In this case the lack of efficient likelihood evaluation precludes the use of Monte Carlo estimators such as SNIS. Diffusion samplers, propose novel objectives for training diffusion models in the absence of an empirical data distribution, both simulation-based~\citep{berner2022optimal,vargas2023denoising,richter2023improved,zhang2021path,vargas2024transport} and simulation-free \citep{akhound2024iterated,huang2021schrodinger,de2024target}. Notably, \citet{havens2025adjointsamplinghighlyscalable} develop a diffusion sampler that is transferable across small molecules. Time coarseners are another family of model, in which ML is used to predict large time transitions for simulation \citep{schreiner2023implicit, fu2023simulate, klein2023timewarp, daigavane2024jamun,yu2025unisimunifiedsimulatortimecoarsened}, whereas methods like MDGen apply generative modeling to both the spatial and temporal dimensions of MD data \citep{jing2024generative}. Lastly, several works integrate normalizing flows with classical Monte Carlo methods \citep{albergo2019flow,arbel2021annealed,gabrie2021efficient, matthews2022continual,midgley2022flow,hagemann2023generalized}.

\section{Conclusion}
\label{sec:conclusion}

We develop \name, demonstrating that deep learning-based samplers can efficiently transfer to previously unseen systems at unprecedented scale. \name outperforms learned baseline methods, as well as molecular dynamics, at a variety of energy evaluation and walltime budgets. Notably, \name demonstrates state-of-the-art performance whilst retaining many simple design choices; thus, leaving many directions for further development. The competitive performance of SNIS compared to SMC invites further investigation into the merits of annealing-based samplers given a proposal with good coverage of the target density. Naturally, annealing-based samplers have been enhanced beyond the simple instantiations we explore; careful tuning of SMC may yield further improvements \citep{syed2024optimised}. We lastly note the self-improvement strategy discussed to not be restricted to SNIS, the integration of advanced Monte Carlo methods presents an avenue for future work.

\paragraph{Limitations} Whilst conventional Monte Carlo algorithms make no assumption on the target density, transferable learned samplers, including \name, rely on the assumption that the system belongs to a structured space of energy functions, in our case the chemical space of peptides. To achieve greater practical relevance it will be necessary to consider a more diverse chemical space, such as the recent OMol25~\citep{levine2025openmolecules2025omol25} dataset. We lastly comment that, despite the scaled prior method demonstrating surprising abilities to sample from the higher temperatures, we believe that precise transfer to lower temperatures would require further algorithmic development.

\clearpage

\section*{Acknowledgments} This research is partially supported by the EP- SRC Turing AI World-Leading Research Fellowship No. EP/X040062/1 and EPSRC AI Hub No. EP/Y028872/1. The authors acknowledge funding from UNIQUE, CIFAR, NSERC, Intel, and Samsung. The research was enabled in part by computational resources provided by
the Digital Research Alliance of Canada (\url{https://alliancecan.ca}), Mila (\url{https://mila.quebec}), and NVIDIA. The authors additionally thank HuggingFace for hosting the ManyPeptidesMD dataset. KN was supported by IVADO and Institut Courtois.

\bibliographystyle{plainnat}
\bibliography{main}

\begin{thebibliography}{75}
\providecommand{\natexlab}[1]{#1}
\providecommand{\url}[1]{\texttt{#1}}
\expandafter\ifx\csname urlstyle\endcsname\relax
  \providecommand{\doi}[1]{doi: #1}\else
  \providecommand{\doi}{doi: \begingroup \urlstyle{rm}\Url}\fi

\bibitem[Abdin and Kim(2024)]{abdin_direct_2024a}
Osama Abdin and Philip~M. Kim.
\newblock Direct conformational sampling from peptide energy landscapes through hypernetwork-conditioned diffusion.
\newblock \emph{Nature Machine Intelligence}, 6\penalty0 (7):\penalty0 775--786, July 2024.
\newblock ISSN 2522-5839.
\newblock \doi{10.1038/s42256-024-00860-4}.

\bibitem[Abramson et~al.(2024)Abramson, Adler, Dunger, Evans, Green, Pritzel, Ronneberger, Willmore, Ballard, Bambrick, Bodenstein, Evans, Hung, O’Neill, Reiman, Tunyasuvunakool, Wu, Žemgulytė, Arvaniti, Beattie, Bertolli, Bridgland, Cherepanov, Congreve, Cowen-Rivers, Cowie, Figurnov, Fuchs, Gladman, Jain, Khan, Low, Perlin, Potapenko, Savy, Singh, Stecula, Thillaisundaram, Tong, Yakneen, Zhong, Zielinski, Žídek, Bapst, Kohli, Jaderberg, Hassabis, and Jumper]{abramson_accurate_2024}
Josh Abramson, Jonas Adler, Jack Dunger, Richard Evans, Tim Green, Alexander Pritzel, Olaf Ronneberger, Lindsay Willmore, Andrew~J. Ballard, Joshua Bambrick, Sebastian~W. Bodenstein, David~A. Evans, Chia-Chun Hung, Michael O’Neill, David Reiman, Kathryn Tunyasuvunakool, Zachary Wu, Akvilė Žemgulytė, Eirini Arvaniti, Charles Beattie, Ottavia Bertolli, Alex Bridgland, Alexey Cherepanov, Miles Congreve, Alexander~I. Cowen-Rivers, Andrew Cowie, Michael Figurnov, Fabian~B. Fuchs, Hannah Gladman, Rishub Jain, Yousuf~A. Khan, Caroline M.~R. Low, Kuba Perlin, Anna Potapenko, Pascal Savy, Sukhdeep Singh, Adrian Stecula, Ashok Thillaisundaram, Catherine Tong, Sergei Yakneen, Ellen~D. Zhong, Michal Zielinski, Augustin Žídek, Victor Bapst, Pushmeet Kohli, Max Jaderberg, Demis Hassabis, and John~M. Jumper.
\newblock Accurate structure prediction of biomolecular interactions with {AlphaFold} 3.
\newblock \emph{Nature}, 2024.

\bibitem[Akhound-Sadegh et~al.(2024)Akhound-Sadegh, Rector-Brooks, Bose, Mittal, Lemos, Liu, Sendera, Ravanbakhsh, Gidel, Bengio, Malkin, and Tong]{akhound2024iterated}
Tara Akhound-Sadegh, Jarrid Rector-Brooks, Joey Bose, Sarthak Mittal, Pablo Lemos, Cheng-Hao Liu, Marcin Sendera, Siamak Ravanbakhsh, Gauthier Gidel, Yoshua Bengio, Nikolay Malkin, and Alexander Tong.
\newblock Iterated denoising energy matching for sampling from boltzmann densities.
\newblock In \emph{International Conference on Machine Learning (ICML)}, 2024.

\bibitem[Albergo and {Vanden-Eijnden}(2023)]{albergo_building_2023}
Michael~S. Albergo and Eric {Vanden-Eijnden}.
\newblock Building normalizing flows with stochastic interpolants.
\newblock \emph{International Conference on Learning Representations (ICLR)}, 2023.

\bibitem[Albergo and Vanden-Eijnden(2025)]{albergo2024nets}
Michael~S Albergo and Eric Vanden-Eijnden.
\newblock Nets: A non-equilibrium transport sampler.
\newblock In \emph{International Conference on Machine Learning (ICML)}, 2025.

\bibitem[Albergo et~al.(2019)Albergo, Kanwar, and Shanahan]{albergo2019flow}
Michael~S Albergo, Gurtej Kanwar, and Phiala~E Shanahan.
\newblock Flow-based generative models for markov chain monte carlo in lattice field theory.
\newblock \emph{Physical Review D}, 2019.

\bibitem[Arbel et~al.(2021)Arbel, Matthews, and Doucet]{arbel2021annealed}
Michael Arbel, Alex Matthews, and Arnaud Doucet.
\newblock Annealed flow transport monte carlo.
\newblock In \emph{International Conference on Machine Learning}, 2021.

\bibitem[Berner et~al.(2024)Berner, Richter, and Ullrich]{berner2022optimal}
Julius Berner, Lorenz Richter, and Karen Ullrich.
\newblock An optimal control perspective on diffusion-based generative modeling.
\newblock \emph{Transactions on Machine Learning Research (TMLR)}, 2024.

\bibitem[Buch et~al.(2011)Buch, Giorgino, and De~Fabritiis]{buch_complete_2011}
Ignasi Buch, Toni Giorgino, and Gianni De~Fabritiis.
\newblock Complete reconstruction of an enzyme-inhibitor binding process by molecular dynamics simulations.
\newblock \emph{Proceedings of the National Academy of Sciences}, 2011.

\bibitem[Case et~al.(2014)Case, Babin, Berryman, Betz, Cai, Cerutti, Cheatham, Darden, Duke, Gohlke, Goetz, Gusarov, Homeyer, Janowski, Kaus, Kolossv{\'a}ry, Kovalenko, Lee, LeGrand, Luchko, Luo, Madej, Merz, Paesani, Roe, Roitberg, Sagui, Salomon-Ferrer, Seabra, Simmerling, Smith, Swails, Walker, Wang, Wolf, Wu, and Kollman]{amber14}
D.~A. Case, V.~Babin, J.~T. Berryman, R.~M. Betz, Q.~Cai, D.~S. Cerutti, T.~E. Cheatham, T.~A. Darden, R.~E. Duke, H.~Gohlke, A.~W. Goetz, S.~Gusarov, N.~Homeyer, P.~Janowski, J.~Kaus, I.~Kolossv{\'a}ry, A.~Kovalenko, T.~S. Lee, S.~LeGrand, T.~Luchko, R.~Luo, B.~Madej, K.~M. Merz, F.~Paesani, D.~R. Roe, A.~Roitberg, C.~Sagui, R.~Salomon-Ferrer, G.~Seabra, C.~L. Simmerling, W.~Smith, J.~Swails, R.~C. Walker, J.~Wang, R.~M. Wolf, X.~Wu, and P.~A. Kollman.
\newblock \emph{AMBER 14}.
\newblock University of California, San Francisco, 2014.

\bibitem[Chen et~al.(2018{\natexlab{a}})Chen, Rubanova, Bettencourt, and Duvenaud]{chen_neural_2018}
Ricky T.~Q. Chen, Yulia Rubanova, Jesse Bettencourt, and David~K Duvenaud.
\newblock Neural ordinary differential equations.
\newblock \emph{Neural Information Processing Systems (NIPS)}, 2018{\natexlab{a}}.

\bibitem[Chen et~al.(2018{\natexlab{b}})Chen, Rubanova, Bettencourt, and Duvenaud]{chen2018neural}
Ricky~TQ Chen, Yulia Rubanova, Jesse Bettencourt, and David~K Duvenaud.
\newblock Neural ordinary differential equations.
\newblock \emph{Advances in neural information processing systems}, 2018{\natexlab{b}}.

\bibitem[Daigavane et~al.(2024)Daigavane, Vani, Saremi, Kleinhenz, and Rackers]{daigavane2024jamun}
Ameya Daigavane, Bodhi~P Vani, Saeed Saremi, Joseph Kleinhenz, and Joshua Rackers.
\newblock Jamun: Transferable molecular conformational ensemble generation with walk-jump sampling.
\newblock \emph{arXiv}, 2024.

\bibitem[De~Bortoli et~al.(2024)De~Bortoli, Hutchinson, Wirnsberger, and Doucet]{de2024target}
Valentin De~Bortoli, Michael Hutchinson, Peter Wirnsberger, and Arnaud Doucet.
\newblock Target score matching.
\newblock \emph{arXiv}, 2024.

\bibitem[Del~Moral(2013)]{del2013mean}
Pierre Del~Moral.
\newblock Mean field simulation for monte carlo integration.
\newblock \emph{Monographs on Statistics and Applied Probability}, 2013.

\bibitem[Dibak et~al.(2022)Dibak, Klein, Kr\"amer, and No\'e]{dibak2021temperature}
Manuel Dibak, Leon Klein, Andreas Kr\"amer, and Frank No\'e.
\newblock Temperature steerable flows and {Boltzmann} generators.
\newblock \emph{Phys. Rev. Res.}, 2022.

\bibitem[Dinh et~al.(2017)Dinh, Sohl-Dickstein, and Bengio]{dinh2016density}
Laurent Dinh, Jascha Sohl-Dickstein, and Samy Bengio.
\newblock Density estimation using {Real NVP}.
\newblock \emph{International Conference on Learning Representations (ICLR)}, 2017.

\bibitem[Doucet et~al.(2001)Doucet, De~Freitas, Gordon, et~al.]{doucet2001sequential}
Arnaud Doucet, Nando De~Freitas, Neil~James Gordon, et~al.
\newblock \emph{Sequential Monte Carlo methods in practice}.
\newblock 2001.

\bibitem[Durkan et~al.(2019)Durkan, Bekasov, Murray, and Papamakarios]{durkan2019neural}
Conor Durkan, Artur Bekasov, Iain Murray, and George Papamakarios.
\newblock Neural spline flows.
\newblock \emph{Advances in neural information processing systems}, 2019.

\bibitem[Eastman et~al.(2017)Eastman, Swails, Chodera, McGibbon, Zhao, Beauchamp, Wang, Simmonett, Harrigan, Stern, Wiewiora, Brooks, and Pande]{eastman_openmm_2017}
Peter Eastman, Jason Swails, John~D. Chodera, Robert~T. McGibbon, Yutong Zhao, Kyle~A. Beauchamp, Lee-Ping Wang, Andrew~C. Simmonett, Matthew~P. Harrigan, Chaya~D. Stern, Rafal~P. Wiewiora, Bernard~R. Brooks, and Vijay~S. Pande.
\newblock {OpenMM} 7: {Rapid} development of high performance algorithms for molecular dynamics.
\newblock \emph{PLoS computational biology}, 2017.

\bibitem[Flamary et~al.(2021)Flamary, Courty, Gramfort, Alaya, Boisbunon, Chambon, Chapel, Corenflos, Fatras, Fournier, Gautheron, Gayraud, Janati, Rakotomamonjy, Redko, Rolet, Schutz, Seguy, Sutherland, Tavenard, Tong, and Vayer]{flamary_pot_2021}
Rémi Flamary, Nicolas Courty, Alexandre Gramfort, Mokhtar~Z. Alaya, Aurélie Boisbunon, Stanislas Chambon, Laetitia Chapel, Adrien Corenflos, Kilian Fatras, Nemo Fournier, Léo Gautheron, Nathalie T.~H. Gayraud, Hicham Janati, Alain Rakotomamonjy, Ievgen Redko, Antoine Rolet, Antony Schutz, Vivien Seguy, Danica~J. Sutherland, Romain Tavenard, Alexander Tong, and Titouan Vayer.
\newblock {POT}: {Python} {Optimal} {Transport}.
\newblock \emph{Journal of Machine Learning Research}, 2021.

\bibitem[Fu et~al.(2023)Fu, Xie, Rebello, Olsen, and Jaakkola]{fu2023simulate}
Xiang Fu, Tian Xie, Nathan~J Rebello, Bradley Olsen, and Tommi~S Jaakkola.
\newblock Simulate time-integrated coarse-grained molecular dynamics with multi-scale graph networks.
\newblock \emph{Transactions on Machine Learning Research}, 2023.

\bibitem[Gabri{\'e} et~al.(2021)Gabri{\'e}, Rotskoff, and Vanden-Eijnden]{gabrie2021efficient}
Marylou Gabri{\'e}, Grant~M Rotskoff, and Eric Vanden-Eijnden.
\newblock Efficient {Bayesian} sampling using normalizing flows to assist {Markov} chain {Monte Carlo} methods.
\newblock \emph{arXiv}, 2021.

\bibitem[Geffner et~al.(2024)Geffner, Didi, Zhang, Reidenbach, Cao, Yim, Geiger, Dallago, Kucukbenli, Vahdat, and Kreis]{geffner_proteina_2024}
Tomas Geffner, Kieran Didi, Zuobai Zhang, Danny Reidenbach, Zhonglin Cao, Jason Yim, Mario Geiger, Christian Dallago, Emine Kucukbenli, Arash Vahdat, and Karsten Kreis.
\newblock Proteina: {Scaling} {Flow}-based {Protein} {Structure} {Generative} {Models}.
\newblock In \emph{International Conference on Learning Representations (ICLR)}, 2024.

\bibitem[Goodfellow et~al.(2014)Goodfellow, Pouget-Abadie, Mirza, Xu, Warde-Farley, Ozair, Courville, and Bengio]{goodfellow2014generativeadversarialnetworks}
Ian~J. Goodfellow, Jean Pouget-Abadie, Mehdi Mirza, Bing Xu, David Warde-Farley, Sherjil Ozair, Aaron Courville, and Yoshua Bengio.
\newblock Generative adversarial networks.
\newblock In \emph{NeurIPS}, 2014.

\bibitem[Grathwohl et~al.(2019)Grathwohl, Chen, Bettencourt, Sutskever, and Duvenaud]{grathwohl_ffjord_2019}
Will Grathwohl, Ricky T.~Q. Chen, Jesse Bettencourt, Ilya Sutskever, and David Duvenaud.
\newblock Ffjord: Free-form continuous dynamics for scalable reversible generative models.
\newblock \emph{International Conference on Learning Representations (ICLR)}, 2019.

\bibitem[Hagemann et~al.(2023)Hagemann, Hertrich, and Steidl]{hagemann2023generalized}
Paul~Lyonel Hagemann, Johannes Hertrich, and Gabriele Steidl.
\newblock \emph{Generalized normalizing flows via Markov chains}.
\newblock 2023.

\bibitem[Havens et~al.(2025)Havens, Miller, Yan, Domingo-Enrich, Sriram, Wood, Levine, Hu, Amos, Karrer, Fu, Liu, and Chen]{havens2025adjointsamplinghighlyscalable}
Aaron Havens, Benjamin~Kurt Miller, Bing Yan, Carles Domingo-Enrich, Anuroop Sriram, Brandon Wood, Daniel Levine, Bin Hu, Brandon Amos, Brian Karrer, Xiang Fu, Guan-Horng Liu, and Ricky T.~Q. Chen.
\newblock Adjoint sampling: Highly scalable diffusion samplers via adjoint matching, 2025.

\bibitem[Ho et~al.(2020)Ho, Jain, and Abbeel]{ho2020denoisingdiffusionprobabilisticmodels}
Jonathan Ho, Ajay Jain, and Pieter Abbeel.
\newblock Denoising diffusion probabilistic models, 2020.

\bibitem[Honda et~al.(2004)Honda, Yamasaki, Sawada, and Morii]{honda200410}
Shinya Honda, Kazuhiko Yamasaki, Yoshito Sawada, and Hisayuki Morii.
\newblock 10 residue folded peptide designed by segment statistics.
\newblock \emph{Structure}, 2004.

\bibitem[Huang et~al.(2021)Huang, Jiao, Kang, Liao, Liu, and Liu]{huang2021schrodinger}
Jian Huang, Yuling Jiao, Lican Kang, Xu~Liao, Jin Liu, and Yanyan Liu.
\newblock Schr{\"o}dinger-{F}{\"o}llmer sampler: sampling without ergodicity.
\newblock \emph{arXiv}, 2021.

\bibitem[Jarzynski(1997)]{jarzynski1997nonequilibrium}
Christopher Jarzynski.
\newblock Nonequilibrium equality for free energy differences.
\newblock \emph{Physical Review Letters}, 1997.

\bibitem[Jing et~al.(2022)Jing, Corso, Chang, Barzilay, and Jaakkola]{jing2022torsional}
Bowen Jing, Gabriele Corso, Jeffrey Chang, Regina Barzilay, and Tommi Jaakkola.
\newblock Torsional diffusion for molecular conformer generation.
\newblock \emph{Advances in Neural Information Processing Systems}, 2022.

\bibitem[Jing et~al.(2024)Jing, St{\"a}rk, Jaakkola, and Berger]{jing2024generative}
Bowen Jing, Hannes St{\"a}rk, Tommi Jaakkola, and Bonnie Berger.
\newblock Generative modeling of molecular dynamics trajectories.
\newblock In \emph{Neural Information Processing Systems (NeurIPS)}, 2024.

\bibitem[Kingma and Dhariwal(2018)]{kingma2018glow}
Durk~P Kingma and Prafulla Dhariwal.
\newblock Glow: Generative flow with invertible 1x1 convolutions.
\newblock \emph{Advances in neural information processing systems}, 2018.

\bibitem[Kingma et~al.(2016)Kingma, Salimans, Jozefowicz, Chen, Sutskever, and Welling]{kingma2016improved}
Durk~P Kingma, Tim Salimans, Rafal Jozefowicz, Xi~Chen, Ilya Sutskever, and Max Welling.
\newblock Improved variational inference with inverse autoregressive flow.
\newblock \emph{Advances in neural information processing systems}, 2016.

\bibitem[Klein and Noe(2024)]{klein_transferable_2024}
Leon Klein and Frank Noe.
\newblock Transferable {Boltzmann} {Generators}.
\newblock 2024.

\bibitem[Klein et~al.(2023{\natexlab{a}})Klein, Foong, Fjelde, Mlodozeniec, Brockschmidt, Nowozin, Noe, and Tomioka]{klein_timewarp_2023}
Leon Klein, Andrew Y.~K. Foong, Tor~Erlend Fjelde, Bruno~Kacper Mlodozeniec, Marc Brockschmidt, Sebastian Nowozin, Frank Noe, and Ryota Tomioka.
\newblock Timewarp: {Transferable} {Acceleration} of {Molecular} {Dynamics} by {Learning} {Time}-{Coarsened} {Dynamics}.
\newblock In \emph{Neural Information Processing Systems (NeurIPS)}, 2023{\natexlab{a}}.

\bibitem[Klein et~al.(2023{\natexlab{b}})Klein, Foong, Fjelde, Mlodozeniec, Brockschmidt, Nowozin, No{\'e}, and Tomioka]{klein2023timewarp}
Leon Klein, Andrew~YK Foong, Tor~Erlend Fjelde, Bruno Mlodozeniec, Marc Brockschmidt, Sebastian Nowozin, Frank No{\'e}, and Ryota Tomioka.
\newblock Timewarp: Transferable acceleration of molecular dynamics by learning time-coarsened dynamics.
\newblock \emph{Neural Information Processing Systems (NeurIPS)}, 2023{\natexlab{b}}.

\bibitem[Klein et~al.(2023{\natexlab{c}})Klein, Krämer, and Noe]{klein_equivariant_2023}
Leon Klein, Andreas Krämer, and Frank Noe.
\newblock Equivariant flow matching.
\newblock \emph{Advances in Neural Information Processing Systems}, 2023{\natexlab{c}}.

\bibitem[Kolesnikov et~al.(2024)Kolesnikov, Pinto, and Tschannen]{kolesnikov2024jet}
Alexander Kolesnikov, Andr{\'e}~Susano Pinto, and Michael Tschannen.
\newblock Jet: A modern transformer-based normalizing flow.
\newblock \emph{arXiv preprint arXiv:2412.15129}, 2024.

\bibitem[Köhler et~al.(2023)Köhler, Invernizzi, Haan, and Noe]{kohler_rigid_2023}
Jonas Köhler, Michele Invernizzi, Pim~De Haan, and Frank Noe.
\newblock Rigid {Body} {Flows} for {Sampling} {Molecular} {Crystal} {Structures}.
\newblock In \emph{Proceedings of the 40th {International} {Conference} on {Machine} {Learning}}, 2023.

\bibitem[Leimkuhler and Matthews(2015)]{leimkuhler2015molecular}
Benedict Leimkuhler and Charles Matthews.
\newblock \emph{Molecular Dynamics: With Deterministic and Stochastic Numerical Methods}.
\newblock 2015.

\bibitem[Levine et~al.(2025)Levine, Shuaibi, Spotte-Smith, Taylor, Hasyim, Michel, Batatia, Csányi, Dzamba, Eastman, Frey, Fu, Gharakhanyan, Krishnapriyan, Rackers, Raja, Rizvi, Rosen, Ulissi, Vargas, Zitnick, Blau, and Wood]{levine2025openmolecules2025omol25}
Daniel~S. Levine, Muhammed Shuaibi, Evan Walter~Clark Spotte-Smith, Michael~G. Taylor, Muhammad~R. Hasyim, Kyle Michel, Ilyes Batatia, Gábor Csányi, Misko Dzamba, Peter Eastman, Nathan~C. Frey, Xiang Fu, Vahe Gharakhanyan, Aditi~S. Krishnapriyan, Joshua~A. Rackers, Sanjeev Raja, Ammar Rizvi, Andrew~S. Rosen, Zachary Ulissi, Santiago Vargas, C.~Lawrence Zitnick, Samuel~M. Blau, and Brandon~M. Wood.
\newblock The open molecules 2025 (omol25) dataset, evaluations, and models, 2025.

\bibitem[Lewis et~al.(2024)Lewis, Hempel, Jim{\'e}nez~Luna, Gastegger, Xie, Foong, Garc{\'\i}a~Satorras, Abdin, Veeling, Zaporozhets, et~al.]{lewis2024scalable}
Sarah Lewis, Tim Hempel, Jos{\'e} Jim{\'e}nez~Luna, Michael Gastegger, Yu~Xie, Andrew~YK Foong, Victor Garc{\'\i}a~Satorras, Osama Abdin, Bastiaan~S Veeling, Iryna Zaporozhets, et~al.
\newblock Scalable emulation of protein equilibrium ensembles with generative deep learning.
\newblock \emph{bioRxiv}, 2024.

\bibitem[Lewis et~al.(2025)Lewis, Hempel, Jim{\'e}nez-Luna, Gastegger, Xie, Foong, Satorras, Abdin, Veeling, Zaporozhets, Chen, Yang, Schneuing, Nigam, Barbero, Stimper, Campbell, Yim, Lienen, Shi, Zheng, Schulz, Munir, Tomioka, Clementi, and No{\'e}]{Lewis2024.12.05.626885}
Sarah Lewis, Tim Hempel, Jos{\'e} Jim{\'e}nez-Luna, Michael Gastegger, Yu~Xie, Andrew Y.~K. Foong, Victor~Garc{\'\i}a Satorras, Osama Abdin, Bastiaan~S. Veeling, Iryna Zaporozhets, Yaoyi Chen, Soojung Yang, Arne Schneuing, Jigyasa Nigam, Federico Barbero, Vincent Stimper, Andrew Campbell, Jason Yim, Marten Lienen, Yu~Shi, Shuxin Zheng, Hannes Schulz, Usman Munir, Ryota Tomioka, Cecilia Clementi, and Frank No{\'e}.
\newblock Scalable emulation of protein equilibrium ensembles with generative deep learning.
\newblock \emph{bioRxiv}, 2025.

\bibitem[Lindorff-Larsen et~al.(2011)Lindorff-Larsen, Piana, Dror, and Shaw]{lindorff-larsen_how_2011}
Kresten Lindorff-Larsen, Stefano Piana, Ron~O. Dror, and David~E. Shaw.
\newblock How {Fast}-{Folding} {Proteins} {Fold}.
\newblock \emph{Science}, 2011.

\bibitem[Liu(2001)]{liu2001monte}
Jun~S Liu.
\newblock \emph{Monte Carlo Strategies in Scientific Computing}.
\newblock 2001.

\bibitem[Liu(2022)]{liu_rectified_2022}
Qiang Liu.
\newblock Rectified flow: A marginal preserving approach to optimal transport.
\newblock \emph{arXiv}, 2022.

\bibitem[Loshchilov and Hutter(2018)]{loshchilov_decoupled_2018}
Ilya Loshchilov and Frank Hutter.
\newblock Decoupled {Weight} {Decay} {Regularization}.
\newblock 2018.

\bibitem[Matthews et~al.(2022)Matthews, Arbel, Rezende, and Doucet]{matthews2022continual}
Alex Matthews, Michael Arbel, Danilo~Jimenez Rezende, and Arnaud Doucet.
\newblock Continual repeated annealed flow transport monte carlo.
\newblock \emph{International Conference on Machine Learning (ICML)}, 2022.

\bibitem[Midgley et~al.(2023{\natexlab{a}})Midgley, Stimper, Antorán, Mathieu, Schölkopf, and Hernández-Lobato]{midgley_se3_2023}
Laurence Midgley, Vincent Stimper, Javier Antorán, Emile Mathieu, Bernhard Schölkopf, and José~Miguel Hernández-Lobato.
\newblock {SE}(3) {Equivariant} {Augmented} {Coupling} {Flows}.
\newblock \emph{Advances in Neural Information Processing Systems}, 2023{\natexlab{a}}.

\bibitem[Midgley et~al.(2023{\natexlab{b}})Midgley, Stimper, Simm, Sch{\"o}lkopf, and Hern{\'a}ndez-Lobato]{midgley2022flow}
Laurence~Illing Midgley, Vincent Stimper, Gregor~NC Simm, Bernhard Sch{\"o}lkopf, and Jos{\'e}~Miguel Hern{\'a}ndez-Lobato.
\newblock Flow annealed importance sampling bootstrap.
\newblock \emph{International Conference on Learning Representations (ICLR)}, 2023{\natexlab{b}}.

\bibitem[Neal(2001)]{neal2001annealed}
Radford~M Neal.
\newblock Annealed importance sampling.
\newblock \emph{Statistics and computing}, 2001.

\bibitem[No{\'e} et~al.(2019)No{\'e}, Olsson, K{\"o}hler, and Wu]{noe2019boltzmann}
Frank No{\'e}, Simon Olsson, Jonas K{\"o}hler, and Hao Wu.
\newblock Boltzmann generators: Sampling equilibrium states of many-body systems with deep learning.
\newblock \emph{Science}, 2019.

\bibitem[Noé et~al.(2009)Noé, Schütte, Vanden-Eijnden, Reich, and Weikl]{noe_constructing_2009}
Frank Noé, Christof Schütte, Eric Vanden-Eijnden, Lothar Reich, and Thomas~R. Weikl.
\newblock Constructing the equilibrium ensemble of folding pathways from short off-equilibrium simulations.
\newblock \emph{Proceedings of the National Academy of Sciences}, 2009.

\bibitem[Papamakarios et~al.(2017)Papamakarios, Pavlakou, and Murray]{papamakarios2017masked}
George Papamakarios, Theo Pavlakou, and Iain Murray.
\newblock Masked autoregressive flow for density estimation.
\newblock In \emph{Neural Information Processing Systems (NeurIPS)}, 2017.

\bibitem[Raja et~al.(2025)Raja, Sipka, Psenka, Kreiman, Pavelka, and Krishnapriyan]{raja2025actionminimization}
Sanjeev Raja, Martin Sipka, Michael Psenka, Tobias Kreiman, Michal Pavelka, and Aditi~S. Krishnapriyan.
\newblock Action-minimization meets generative modeling: Efficient transition path sampling with the onsager-machlup functional.
\newblock In \emph{Forty-second International Conference on Machine Learning}, 2025.
\newblock URL \url{https://openreview.net/forum?id=QwoGfQzuMa}.

\bibitem[Rezende and Mohamed(2015)]{rezende2015variational}
Danilo Rezende and Shakir Mohamed.
\newblock Variational inference with normalizing flows.
\newblock \emph{International Conference on Machine Learning (ICML)}, 2015.

\bibitem[Richter et~al.(2024)Richter, Berner, and Liu]{richter2023improved}
Lorenz Richter, Julius Berner, and Guan-Horng Liu.
\newblock Improved sampling via learned diffusions.
\newblock \emph{International Conference on Learning Representations (ICLR)}, 2024.

\bibitem[Roberts and Tweedie(1996)]{roberts1996exponential}
Gareth~O Roberts and Richard~L Tweedie.
\newblock Exponential convergence of langevin distributions and their discrete approximations.
\newblock \emph{Bernoulli}, 1996.

\bibitem[Satorras et~al.(2021)Satorras, Hoogeboom, and Welling]{satorras2021n}
V{\i}ctor~Garcia Satorras, Emiel Hoogeboom, and Max Welling.
\newblock E (n) equivariant graph neural networks.
\newblock \emph{International Conference on Machine Learning (ICML)}, 2021.

\bibitem[Schopmans and Friederich(2025)]{schopmans_temperature-annealed_2025}
Henrik Schopmans and Pascal Friederich.
\newblock Temperature-{Annealed} {Boltzmann} {Generators}, 2025.

\bibitem[Schreiner et~al.(2023)Schreiner, Winther, and Olsson]{schreiner2023implicit}
Mathias Schreiner, Ole Winther, and Simon Olsson.
\newblock Implicit transfer operator learning: Multiple time-resolution models for molecular dynamics.
\newblock In \emph{Thirty-seventh Conference on Neural Information Processing Systems}, 2023.

\bibitem[Shazeer(2020)]{shazeer2020gluvariantsimprovetransformer}
Noam Shazeer.
\newblock Glu variants improve transformer, 2020.
\newblock URL \url{https://arxiv.org/abs/2002.05202}.

\bibitem[Song et~al.(2021)Song, Sohl-Dickstein, Kingma, Kumar, Ermon, and Poole]{song2021scorebasedgenerativemodelingstochastic}
Yang Song, Jascha Sohl-Dickstein, Diederik~P. Kingma, Abhishek Kumar, Stefano Ermon, and Ben Poole.
\newblock Score-based generative modeling through stochastic differential equations, 2021.

\bibitem[Syed et~al.(2024)Syed, Bouchard-C{\^o}t{\'e}, Chern, and Doucet]{syed2024optimised}
Saifuddin Syed, Alexandre Bouchard-C{\^o}t{\'e}, Kevin Chern, and Arnaud Doucet.
\newblock Optimised annealed sequential monte carlo samplers.
\newblock \emph{arXiv preprint arXiv:2408.12057}, 2024.

\bibitem[Tan et~al.(2025)Tan, Bose, Lin, Klein, Bronstein, and Tong]{tan_scalable_2025}
Charlie~B. Tan, Avishek~Joey Bose, Chen Lin, Leon Klein, Michael~M. Bronstein, and Alexander Tong.
\newblock Scalable {Equilibrium} {Sampling} with {Sequential} {Boltzmann} {Generators}, 2025.

\bibitem[Vargas et~al.(2023)Vargas, Grathwohl, and Doucet]{vargas2023denoising}
Francisco Vargas, Will Grathwohl, and Arnaud Doucet.
\newblock Denoising diffusion samplers.
\newblock \emph{International Conference on Learning Representations (ICLR)}, 2023.

\bibitem[Vargas et~al.(2024)Vargas, Padhy, Blessing, and N{\"u}sken]{vargas2024transport}
Francisco Vargas, Shreyas Padhy, Denis Blessing, and Nikolas N{\"u}sken.
\newblock Transport meets variational inference: Controlled {Monte Carlo} diffusions.
\newblock \emph{International Conference on Learning Representations (ICLR)}, 2024.

\bibitem[Wayment-Steele et~al.(2024)Wayment-Steele, Ojoawo, Otten, et~al.]{waymentSteele2024}
Hannah~K. Wayment-Steele, Adebayo Ojoawo, Ren{\'e} Otten, et~al.
\newblock Predicting multiple conformations via sequence clustering and alphafold2.
\newblock \emph{Nature}, 2024.

\bibitem[Wu and Brooks(2004)]{wu_beta-hairpin_2004}
Xiongwu Wu and Bernard~R. Brooks.
\newblock Beta-hairpin folding mechanism of a nine-residue peptide revealed from molecular dynamics simulations in explicit water.
\newblock \emph{Biophysical Journal}, 2004.

\bibitem[Yu et~al.(2025)Yu, Huang, and Liu]{yu2025unisimunifiedsimulatortimecoarsened}
Ziyang Yu, Wenbing Huang, and Yang Liu.
\newblock Unisim: A unified simulator for time-coarsened dynamics of biomolecules, 2025.

\bibitem[Zhai et~al.(2024)Zhai, Zhang, Nakkiran, Berthelot, Gu, Zheng, Chen, Bautista, Jaitly, and Susskind]{zhai_normalizing_2024}
Shuangfei Zhai, Ruixiang Zhang, Preetum Nakkiran, David Berthelot, Jiatao Gu, Huangjie Zheng, Tianrong Chen, Miguel~Angel Bautista, Navdeep Jaitly, and Josh Susskind.
\newblock Normalizing {Flows} are {Capable} {Generative} {Models}, 2024.

\bibitem[Zhang and Chen(2022)]{zhang2021path}
Qinsheng Zhang and Yongxin Chen.
\newblock Path integral sampler: a stochastic control approach for sampling.
\newblock \emph{International Conference on Learning Representations (ICLR)}, 2022.

\end{thebibliography}

\clearpage
\appendix

\renewcommand \thepart{}
\renewcommand \partname{}

\part{Appendix}\label{appendix} %

\startcontents[appendix]
\printcontents[appendix]{l}{1}

\clearpage

\section{\name architecture Details}
\label{app:architecture}

\subsection{Tokenization and masking} 

As in SBG \citep{tan_scalable_2025}, the token sequence is constructed using a single atom per token $x[i]\in\mathbb{R}^3$. As different peptide sequences contain varying numbers of atoms, we zero-pad atom sequences to a fixed maximum sequence length and introduce a padding mask $m[i]\in{0,1}$ indicating valid ($m[i]=1$) or padded ($m[i]=0$) tokens. In the context of a causal transformer, the implementation is greatly simplified by ensuring all padding tokens are placed at the end of the token sequence, irrespective of the permutations applied. We may therefore state the following block update rules, as only a minor adaptation of \cref{eq:tarflow_fwd}
\begin{align}
z_{t+1}[i] =
\begin{cases}
z_t[i], & i=0, \\
(z_t[i]-\mu_t(z_t[:i])[i])\odot\exp(-\alpha_t(z_t[:i])[i]), & m[i] = 1, \\
[0,0,0], & m[i] = 0, \\
\end{cases}
\end{align}
with log-determinant of Jacobian given by
\begin{align}
\log\left|\deriv{f_t(z_{t})}{z_{t}}\right| = -\sum_{i=1}^{N-1}\sum_{j=0}^{D-1} m[i] \cdot \alpha_t(z_{t}[:i])[i]_j.
\end{align}
The inverse transformation is correspondingly a minor adaptation of \cref{eq:tarflow_rev}
\begin{align}
z_{t}[i] =
\begin{cases}
z_{t+1}[i], & i=0, \\
z_{t+1}[i] \odot \exp(\alpha_t(z_t[:i])[i]) + \mu_t(z_t[:i])[i], & m[i] = 1, \\
[0,0,0], & m[i] = 0. \\
\end{cases}
\end{align}

\subsection{Permutations}

All permutations are defined from the N-terminal to the C-terminal. In the residue-by-residue permutation the atoms are ordered such that each residue forms a contiguous sequence, with sidechain atoms immediately following the corresponding backbone atoms. In the backbone-first permutation the entire sequence of backbone atoms is placed at the start of the sequence before any sidechain atoms. Where constituent residues possess a branch or ring we also introduce a variant in which the branch ordering is flipped or ring traversal is inverted. The flip permutations are a simple inversion of the permutation, where padding tokens and not moved from their position at the end of the sequence. The specific sequence of permutations employed in \name is presented in \cref{tab:perm_order}.

\begin{table}[h]
\centering
\caption{Autoregressive permutation order used across the eight transformation blocks.}
\label{tab:perm_order}
\begin{tabular}{ll}
\toprule
Permutation & Description \\
\midrule
$\pi_0$ & Backbone first \\
$\pi_1$ & Residue–by–residue (flip) \\
$\pi_2$ & Backbone–first (flip) \\
$\pi_3$ & Residue–by–residue \\
$\pi_4$ & Backbone–first (variant) \\
$\pi_5$ & Residue–by–residue (variant, flip) \\
$\pi_6$ & Backbone–first (variant, flip) \\
$\pi_7$ & Residue–by–residue (variant) \\
\bottomrule
\end{tabular}
\end{table}

\subsection{Adaptive layer norm and transition}

\name integrates the adaptive layer normalization and SwiGLU-based \citep{shazeer2020gluvariantsimprovetransformer} transition modules employed by \citet{geffner_proteina_2024} into the transformer blocks of the TarFlow architecture \citep{zhai_normalizing_2024}. 
The positions of the latent vector \(z_t\) are encoded using a sinusoidal positional encoding, which are added directly to \(z_t\).
The conditional embedding is used in the adaptive layer normalization and adaptive scale components.

\begin{figure}[ht!]
    \centering
    \includegraphics[width=\linewidth, trim=0cm 8cm 0cm 8cm, clip]{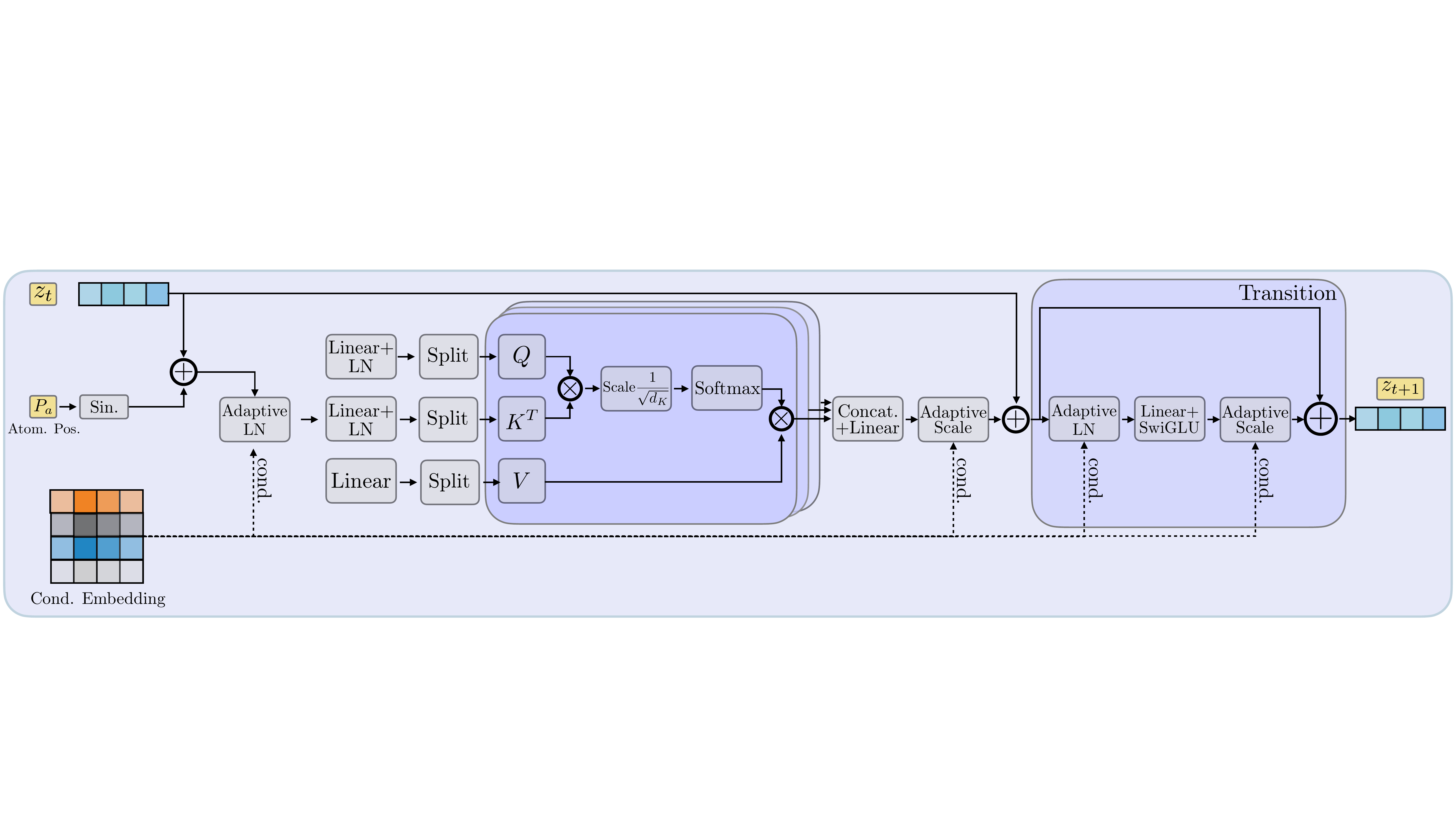}
    \caption{\textbf{Adaptive Layer Norm and Transition.} The transformer block is modified to incorporate conditional information using adaptive layer normalization and a transition block. Figure adapted from \citet{geffner_proteina_2024}.}
    \label{fig:adapt-ln}
\end{figure}

\section{Dataset}
\label{app:dataset}

\paragraph{Sequence sampling} Training sequences are collected for all peptide lengths by uniformly sampling the 20 standard amino acids. For the 8-residue test data, a sequence of length \(30 \cdot 8 = 240\) is constructed by concatenating \(12\) of each amino acid. This sequence is then randomly permuted and split into peptides of length 8, ensuring that each amino acid is represented uniformly. A similar process is performed for length 4 but was not possible at length 2. In both training and test sets, the N- and C-terminal residues are protonated to form the zwitterionic state of the peptides. Initial structure files (PDB format) are generated using AmberTools' \texttt{tleap}.

\paragraph{Molecular dynamics simulation} Local energy minimization is performed with the Limited-memory Broyden–Fletcher–Goldfarb–Shanno (L-BGFS) algorithm. Energy minimization is followed by burn-in simulation of length \SI{50}{\ps}, after which samples are collected every \SI{1}{\ps} (train) or \SI{10}{\ps} (test) until the simulation budget is exhausted. Full MD simulation parameters are provided in \cref{tab:md_parameters}.

\begin{table}[h]
\centering
\caption{\texttt{OpenMM} simulation parameters.}
\label{tab:md_parameters}
\begin{tabular}{ll}
\toprule
Force field         & Amber14 \\
Integration time step            & 1 fs \\
Friction coefficient & \SI{0.3}{\ps}\(^{-1}\) \\
Temperature          & \SI{310}{\K} \\
Nonbonded method      & \texttt{CutoffNonPeriodic} \\
Nonbonded cutoff      & \SI{2}{\nm} \\
Integrator           & \texttt{LangevinMiddleIntegrator} \\
\bottomrule
\end{tabular}
\end{table}

\begin{table}[h]
\centering
\caption{Training and evaluation dataset parameters.}
\label{tab:dataset_parameters}
\begin{tabular}{lll}
\toprule
 & Train & Test \\
 \midrule
Burn-in period & \SI{50}{\ps} & \SI{50}{\ps} \\
Sampling interval & \SI{1}{\ps} & \SI{10}{\ps} \\
Simulation time & \SI{200}{\ns} & \SI{5}{\us} \\
\bottomrule
\end{tabular}
\end{table}

\section{Training configuration}
\label{app:training_details}

All models are trained for \sci{5}{5} iterations using a batch size of 512 with the AdamW optimizer \citep{loshchilov_decoupled_2018}. We employ a cosine learning rate schedule in which the initial and final learning rates are a reduction of the maximal value by factor of 500, as well as exponential moving average with decay of 0.999. No overfitting was observed hence no early stopping was required. Given the large size of the training trajectories, we subsample to \SI{10}{\ps} per frame. Samples are normalized using values approximating the standard deviation of the 8AA data ($\sigma = 0.35$), or if absent the 4AA data ($\sigma =0.28$), noting that a single value must be shared across systems of different dimensionality. An overview of all training configurations is provided in \cref{tab:training_config}.

\paragraph{Continuous Normalizing Flows} We use the ECNF++ training recipe defined by \citet{tan_scalable_2025}; this entails a learning rate of \sci{5}{-4} and weight decay of \sci{1}{-4}, with default AdamW hyperparameters of AdamW \(\beta_1,\beta_2\) of \((0.9, 0.999)\). In contrast the ECNF of \citet{klein_transferable_2024} was trained without weight decay or exponential weight averaging. The channel width and layer depth of both models is defined in \cref{tab:scaling_parameters}.

\paragraph{TarFlows} Following \citet{zhai_normalizing_2024} and \citet{tan_scalable_2025} we use a learning rate of \sci{1}{-4}, weight decay of \sci{4}{-4}, and AdamW \(\beta_1,\beta_2\) of \((0.9, 0.95)\). Data augmentation is applied as random rotations and Gaussian center of mass augmentation, in which every the entire system conformation is translated by a vector \(c \sim \mathcal{N}(0,\sigma_c^2I_3)\). The \(\sigma^2\) value is chosen to match that of the prior, which has a center of mass variance \(\sigma_c^2 = \frac{1}{N}\) where \(N\) is the number of atoms. Given \(N\) is in our case variable for a single model trained on multiple systems, this augmentation is applied-per system before padding is applied. The architecture width and depth is provided in \cref{tab:scaling_parameters}.

\begin{table}[h!]
    \centering
    \caption{Overview of training configurations.}
    \label{tab:training_config}
    \begin{tabular}{l l l l}
        \toprule
         & ECNF & ECNF++ & TarFlow / \name\\
        \midrule
        Learning Rate & \sci{5}{-4} & \sci{5}{-4} & \sci{1}{-4} \\
        Weight Decay & 0.0 & \sci{1}{-2} & \sci{4}{-4} \\
        $\beta_1,\beta_2$ & 0.9, 0.999 & 0.9, 0.999 & 0.9, 0.95\\
        EMA Decay & 0.0 & 0.999 & 0.999\\
        \bottomrule
    \end{tabular}
\end{table}

\begin{table}[h!]
    \centering
    \caption{Overview of model scaling parameters. For TarFlow variants depth corresponds to number of parameterized transformations, for ECNF variants this is simply the number of graph neural network layers.}
    \label{tab:scaling_parameters}
    \begin{tabular}{l l l l l}
        \toprule
         & ECNF & ECNF++ & TarFlow & \name\\
        \midrule
        Channels & 128 & 256 & 384 & 384 \\
        Depth & 9 & 9 & 8 & 8\\
        Layers per block & N/A & N/A & 8 & 8\\
        Parameters (M) & 1 & 4 & 115 & 285 \\
        \bottomrule
    \end{tabular}
\end{table}

\subsection{Computational resources}

All training experiments are run NVIDIA H100 GPUs using distributed data parallelism. The training throughput for each model is presented in \cref{tab:training_throughput}.

\begin{table}[h!]
    \centering
    \caption{Training throughput for models presented in \cref{tab:main_results}. We highlight ECNF++ to be trained only on sequences up to length 4, whereas TarFlow and \name are trained on sequences up to length 8.}
    \label{tab:training_throughput}
    \begin{tabular}{l l l l l}
        \toprule
         & ECNF++ & TarFlow & \name\\
        \midrule
         Training iterations / H100 hour & 960 & 1132 & 260 \\
        \bottomrule
    \end{tabular}
\end{table}

\section{Importance sampling variants}
\label{app:importance_sampling}

\subsection{Self-normalized importance sampling}

Self-normalized importance sampling (SNIS) corresponds to the following estimator
\begin{align}
    \mean_{p(x)}\varphi(x) \approx \sum_{i=1}^n \frac{w_i}{\sum_{j=1}^n w_j}\varphi(x_i)\,,\;\; w_i = \frac{p(x_i)}{q(x_i)}\,,\;\; x_i \sim q(x)\,,
\end{align}
where one uses the learned density model of \name $q_\theta(x)$ as $q(x)$.

\subsection{Continuous-time annealed importance sampling}
\label{app:cont-time-ais}

Below, we repeat the derivations from \citep{jarzynski1997nonequilibrium, albergo2024nets}. Namely, we consider the continuous family of marginal densities
\begin{align}
    q_t(x) =\frac{1}{Z_t} \exp\left(-E_t(x)\right)\,, \;\; Z_t = \int dx\;\exp\left(-E_t(x)\right)\,.
\end{align}
The PDE describing the time-evolution of this density is
\begin{align}
    \deriv{q_t(x)}{t} =~& q_t(x)\left[-\deriv{E_t(x)}{t} + \mean_{q_t(x)}\deriv{E_t(x)}{t}\right]\\
    =~& \pm \inner{\nabla}{q_t(x)\frac{\sigma_t^2}{2}\nabla\log q_t(x)} + q_t(x)\left[-\deriv{E_t(x)}{t} + \mean_{q_t(x)}\deriv{E_t(x)}{t}\right]\\
    =~& - \inner{\nabla}{q_t(x)\frac{\sigma_t^2}{2}\nabla\log q_t(x)} + \frac{\sigma_t^2}{2}\Delta q_t(x) + q_t(x)\left[-\deriv{E_t(x)}{t} + \mean_{q_t(x)}\deriv{E_t(x)}{t}\right]\,.
\end{align}
This is a Feynman-Kac PDE which can be simulated \citep{del2013mean} as the following SDE on the extended space of states $x_t$ and weights $w_t$
\begin{align}
    \begin{split}
    dx_t =~& -\frac{\sigma_t^2}{2}\nabla E_t(x)dt + \sigma_tdW_t\,, \;\; x_{t=0} \sim q_0(x)\\
    d\log w_t =~& -\deriv{E_t(x)}{t}dt\,,\;\; w_{t=0} = 1\,.    
    \label{appeq:wsde}
    \end{split}
\end{align}
The expectation of the statistics $\varphi(x)$ w.r.t. the density $q_T(x)$ then can be estimated using SNIS as follows
\begin{align}
    \mean_{q_T(x)}\varphi(x) \approx \sum_{i=1}^n \frac{w_T^i}{\sum_{j=1}^n w_T^j}\varphi(x_T^i)\,,
\end{align}
where $(x_T^i, w_T^i)$ are the solutions of the SDE \cref{appeq:wsde}.

For the inference time of \name, we define the continuous family of marginals as
\begin{align}
    q_t(x) \propto \exp\big(\underbrace{(1-t)\log q_\theta(x) + t\log p(x)}_{-E_t(x)}\big)\,,\;\; t \in [0,1]\,,
\end{align}
where $q_\theta(x)$ is the learned density of \name. Thus, \cref{appeq:wsde} becomes
\begin{align}
    \begin{split}
    dx_t =~& \frac{\sigma_t^2}{2}\left((1-t)\nabla\log q_\theta(x) + t\nabla\log p(x)\right)dt + \sigma_tdW_t\,, \;\; x_{t=0} \sim q_0(x)\\
    d\log w_t =~& \left(\log p(x) - \log q_\theta(x)\right)dt\,,\;\; w_{t=0} = 1\,.    
    \end{split}
\end{align}

\subsection{Discrete-time annealed importance sampling}
\label{app:discrete-time-ais}

Consider a sequence of marginal densities
\begin{align}
    q_0(x)\propto \exp(-E_0(x)),\ldots,q_K(x)\propto \exp(-E_K(x))\,.
\end{align}
Let's denote by $k_t(x_t\cond x_{t-1})$ the kernel that satisfies the detailed balance w.r.t. $q_t(x_t)$, i.e.
\begin{align}
    q_t(x_{t-1})k_t(x_t\cond x_{t-1}) = q_t(x_{t})k_t(x_{t-1}\cond x_t)\,.
\end{align}
Then, one can write importance sampling estimator for the final marginal as
\begin{align}
    \int dx_K\; q_K(x_K)\varphi(x_K) =~& \int dx_{K-1}dx_K\; k_K(x_{K-1}\cond x_K)q_K(x_K) \varphi(x_K) \\
    =~& \int dx_{K-1}dx_K\; k_K(x_{K}\cond x_{K-1})q_K(x_{K-1}) \varphi(x_K) \\
    =~& \mean_{q_{K-1}(x_{K-1})k_K(x_{K}\cond x_{K-1})}\frac{q_K(x_{K-1})}{q_{K-1}(x_{K-1})} \varphi(x_K)\,.
\end{align}
Clearly, we can repeat the trick but now for $q_{K-1}(x_{K-1})$. Thus, applying this trick recursively to different marginals, we have
\begin{align}
    \int dx_K\; q_K(x_K)\varphi(x_K) = \mean_{q_{0}(x_{0})}\mean_{x_1,\ldots,x_K}\prod_{t=0}^{K-1}\frac{q_{t+1}(x_t)}{q_t(x_t)} \varphi(x_{K})\,,\\
    \text{ where } x_1,\ldots,x_K \sim k_1(x_1\cond x_0)\ldots k_{K-1}(x_{K-1}\cond x_{K-2})k_K(x_K\cond x_{K-1})\,.
\end{align}
Thus, we have the following SNIS estimator
\begin{align}
    \int dx\; q_K(x)\varphi(x) \approx~& \sum_i \frac{w_K^i}{\sum_j w_K^j}\varphi(x^i_K)\,,\\
     \text{ where }~&x^i_t \sim k_t(x_t\cond x_{t-1})\,,\;\; t=1,\ldots,K\,,\;\; x_0 \sim q_0(x_0)\\
    ~&\log w_t^i = -E_t(x_{t-1}) + E_{t-1}(x_{t-1}) + \log w_{t-1}^i\,.
\end{align}
Note that there is a lot of flexibility for the choice of $k_t(x_t\cond x_{t-1})$ because we do not use the densities of the transition kernel for the weights. In particular, the Metropolis-Hastings algorithm with any proposal yields a reversible kernel (satisfies the detailed balance), which result in a consistent final estimator. Furthermore, compared to the continuous-time AIS, discrete-time AIS does not introduce the time-discretization error.

At the inference step of \name, we choose
\begin{align}
    q_t(x) \propto \exp\big(\underbrace{(1-t)\log q_\theta(x) + t\log p(x)}_{-E_t(x)}\big)\,,\;\; t = 0,\frac{1}{K},\frac{2}{K},\ldots,1\,,
\end{align}
and Metropolis-Adjusted Langevin Dynamics as the transition kernel $k_t(x_t\cond x_{t-1})$ \citep{roberts1996exponential}.

\subsection{Sequential Monte Carlo}

Sequential Monte Carlo (SMC) \citep{doucet2001sequential} can be understood as
annealed importance sampling equipped with \emph{adaptive resampling}. As in AIS,
particles and weights are propagated in a coupled system. However, SMC tracks the
effective sample size (ESS)
\begin{align}
    \text{ESS}(w^1,\ldots,w^N)
    = \frac{\left(\sum_{i=1}^N w^i\right)^2}{\sum_{i=1}^n (w^i)^2}\,,
\end{align}
and performs a resampling step whenever ESS falls below a chosen threshold. This is in contrast to AIS, in which resampling only occurs in the final timestep.
This adaptive resampling seeks to mitigate weight degeneracy, in which large compute allocations are required in AIS to propogate very low weight particles. While the continuous-time and discrete-time AIS variants are both suitable for SMC, only the continuous-time variant was considered by \citet{tan_scalable_2025}.

\section{Evaluation configuration}
\label{app:evaluation_details}

\subsection{Proposal sampling and likelihood evaluation}

\paragraph{Equivariant continuous normalizing flows} Sampling from a continuous normalizing flow (CNF) involves solving the ODE defined by the parameterized vector field \(u_{t}:[0,1] \times \R^{n \times 3} \to \R^{n \times 3}\) 

\begin{equation}
    \frac{dx_t}{dt} = u_t(x_t), \quad x_0 \sim p_0
    \label{eq:cnf_ode}
\end{equation}

The corresponding likelihoods can be obtained using the instantaneous change of variables formula

\begin{equation}
    \log p_1 (x_1) = \log p_0(x_0) - \int^1_0 \nabla \cdot u_t (x_t) dt,
    \label{eq:instantaneous_change_of_variable}
\end{equation}

where \(\nabla \cdot\) is the divergence operator. In practice both \cref{eq:cnf_ode} and \cref{eq:instantaneous_change_of_variable} can be integrated simultaneously with an ordinary differential equation (ODE) solver. We use the Dormand–Prince-5 (dopri5) adaptive solver in all ECNF experiments. Given the \(E(3)\) equivariance of the ECNF, samples are generated in both possible global chiralities. Following \citet{klein_transferable_2024} we check for incorrect global sample chirality and flip samples appropriately to match the L-amino acids present in the evaluation data. Unlike \citet{klein_transferable_2024} we do not omit any samples with unresolvable chirality. We additionally apply logit clipping, removing the samples with the 0.2\% highest importance weights before resampling \citet{midgley_se3_2023}.

\paragraph{TarFlow variants} As discussed in \cref{sec:nf}, samples are generated from a normalizing flow simply by applying \(f_\theta\) to prior samples \(z \sim \mathcal{N}(0, I_{N\times D})\). Model likelihoods are obtained using the change of variables formula (\cref{eq:change_of_variables}). Given the lack of translation equivariance in the TarFlow, and the data augmentation applied during training, samples are generated with an approximate scaled \(\chi_3\) distribution over centroid norm \(||c|| = ||\frac{1}{N}\sum_{i=1}^N x_{i,:}|| \sim \sigma \chi_3\). This leads to adverse behavior when resampling with finite samples, hence we apply the center of mass adjustment of \citep{tan_scalable_2025}, in which the \(\chi_3\) probability density function is divided out of the proposal likelihoods

\begin{equation}
    \log p^c_{\theta}(x) = \log p_{\theta}(x) - \left[\log\left(\frac{\| c\|^2}{\sigma^3}\right) + \frac{\| c \|^2}{2\sigma^2}  - \log\left(\sqrt{2}\Gamma\left(\frac{3}{2}\right)\right)\right]
    \label{eq:com_adjust}
\end{equation}
where  \(\Gamma\) is the gamma function. This adjustment seeks to account for the radial component introduced by translation non-equivariance. We additionally apply the same weight clipping threshold as in the ECNF when performing SNIS, or before SMC.

\subsection{Metrics}\label{app:metrics}

We report both effective sample size and a variety of Wasserstein-2 distances as evaluation metrics. For the Wasserstein distances a subsample of \scione{4} samples are randomly sampled from the evaluation trajectory as ground truth. Similarly, at most \scione{4} generated samples are employed; if a method has generated more samples a random subset is drawn without replacement.

\paragraph{Effective sample size} We compute the effective sample size (ESS) using Kish's formula, normalized by the number of samples generated

\begin{equation}
\text{ESS}\left(\{w_i\}_{i=1}^N\right) = \frac{\left( \sum_{i=1}^{N} w_i \right)^2}{N\sum_{i=1}^{N} w_i^2}.
\end{equation}

\paragraph{Empirical Wasserstein distance} We compare generated samples to ground truth data, collected as defined in \cref{app:dataset}, using empirical Wasserstein-2 distances. Given empirical distributions \(\mu = \frac{1}{n} \sum_{i=1}^n \delta_{x_i}\) and \(\nu = \frac{1}{m} \sum_{j=1}^m \delta_{y_j}\), the empirical Wasserstein-2 distance is defined as

\begin{equation}
W_2(\mu, \nu) = \min_{\pi \in \Pi(\mu, \nu)} \sqrt{\sum_{i=1}^{n} \sum_{j=1}^{m} \pi_{ij} \, c(x_i, y_j)^2}
\end{equation}

where \(\Pi(\mu, \nu)\) denotes the set of couplings with marginals \(\mu\) and \(\nu\), and \(c(x,y)^2\) is a defined cost function. Different choices of \(c(x,y)^2\) define different measures of dissimilarity. We use the \texttt{POT} \citep{flamary_pot_2021} linear optimal transport solver to compute the optimal couplings.

\paragraph{Energy cost} The energy of a sample \(E(x)\) is sensitive to both bonded forces and non-bonded forces. For the energy Wasserstein-2 distance \emetric the cost function is simply

\begin{equation}
c_E(x,y)^2 = \left|E(x) - E(y)\right|^2
\end{equation}

\paragraph{Dihedral torus cost}
The \(\phi\) and \(\psi\) backbone dihedral angles of a peptide conformation encode essential information regarding secondary and tertiary structure. We compare generated and ground truth samples in angle space by defining the dihedral angle vector

\begin{equation}
\operatorname{Dihedrals}(x) = (\phi_1, \psi_1, \phi_2, \psi_2, \ldots, \phi_{L-1}, \psi_{L-1})
\end{equation}

where \(L\) is the number of residues. Given the torus geometry implied by angle periodicity \(\phi_i \in (-\pi,\pi]\), a natural cost function is the minimal signed angle difference 

\begin{equation}
c_\mathcal{T}(x, y)^2 = \sum_{i=1}^{2 L} \left[ (\operatorname{Dihedrals}(x)_i - \operatorname{Dihedrals}(y)_i + \pi) \bmod 2\pi - \pi \right]^2.
\end{equation}

This metric captures the geometric dissimilarity in dihedral angle space, respecting periodicity.

\paragraph{Time-lagged independent component analysis cost} The time-lagged independent component analysis (TICA) projection of time-series data captures directions along which the data exhibits maximal autocorrelation. Within molecular dynamics, TICA is commonly used to detect distinct metastable states. Given mean-free time series data \(\tilde{x}_t\), the instantaneous (zero-lag) empirical covariance and time-lagged empirical covariance matrix (at lag time \(\tau\)) are computed as

\begin{equation}
\hat{C}_{00} = \frac{1}{T-\tau}\sum_{t=1}^{T-\tau} \tilde{x}_t \tilde{x}_t^\top, \quad \hat{C}_{0\tau} = \frac{1}{T-\tau}\sum_{t=1}^{T-\tau} \tilde{x}_t \tilde{x}_{t+\tau}^\top.
\end{equation}

TICA seeks linear projection vectors \(w\ \in \mathbb{R}^n\) that maximize autocorrelation at lag \(\tau\)

\begin{equation}
\max_{w} \frac{w^\top \hat{C}_{0\tau} w}{w^\top \hat{C}_{00} w}.
\end{equation}

The solution to which is obtained by solving the generalized eigenvalue problem

\begin{equation}
C_{0\tau} \lambda = \lambda C_{00} w,
\label{eq:tica_eigenvalue}
\end{equation}

where the eigenvalue $\lambda$ measures the autocorrelation of the projected component at lag $\tau$, and the eigenvector \(w\) defines the corresponding slow mode. To define the TICA Wasserstein-2 distance \ticametric we take the full evaluation trajectory \emph{without subsampling} and solve \cref{eq:tica_eigenvalue} to obtain the first two TICA projection vectors \(w_1, w_2\). We may then define the following cost function 

\begin{equation}
c_\text{TICA}(x,y)^2 = \sum_{j=1}^2\left[w_j^\top x - w_j^\top y\right]^2
\end{equation}

defining similarity in TICA projection space. In practice, we compute the TICA projection for the heavy-atom pairwise distances and dihedral angles. We also emphasize that the TICA projection must be computed on the full \SI{5}{\us} evaluation trajectory (such that the slowest transitions may be detected), but that the samples \(y\) used in the Wasserstein metric are restricted to the \scione{4} subset.

\paragraph{Jenson-Shannon divergence metrics} Following \citet{raja2025actionminimization}, on the reference MD trajectory we run $k$-means ($k=20$) on the first two time-lagged independent components (TICs) fitted to the pairwise distances and dihedral angles of the peptide confirmations. Using these clusters, we obtain the occupancy distributions for the reference MD and the samples generated from \name, and report their Jenson-Shannon divergence (\ticajsd), defined as
\begin{equation}
    \text{JSD}(P, Q) = \frac{1}{2}\text{KL}(P, M) + \frac{1}{2}\text{KL}(Q, M), \; \text{where}\; \text{KL}(P,M)=\sum_i P_i \log \frac{P_i}{M_i}
\end{equation}
where $P$ and $Q$ are discrete distributions, and $M= \frac{1}{2}(P + Q)$. Moreover, we repeat the same procedure using $k$-means only on the dihedral angles, and report this as \torusjsd. In both cases we fit the $k$-means clustering on the features (TICA projections or dihedrals) from full reference MD, but compute the metrics of generated samples against a subsampling of the reference trajectory as previously discussed. Note that JSD depends on the arbitrary choice of clustering (here $k=20$) and can be sensitive to binning resolution, whereas Wasserstein distances avoid discretization errors by operating directly on the continuous distributions.

\subsection{Sampling algorithm configurations}

In this section we define configurations for the sampling algorithms presented in \cref{tab:sampling_alg_results}, as well as the molecular dynamics baseline used in \cref{fig:usvsmd} and \cref{fig:sael_tica}. In particular the allocation of the \scione{6} energy evaluations within the method is defined.

\paragraph{Molecular dynamics baseline} We follow the same procedure used for collecting the main datasets defined in \cref{app:dataset}, where the parameters defined in \cref{tab:md_parameters} are unchanged. We apply a logarithmically decaying frame interval with appropriate reweighting to obtain accurate resolution across many orders of magnitude. The simulation is run for (\SI{1}{\us}) using \(10^9\) energy evaluations.

\paragraph{Sequential Monte Carlo} For both variants we generate a proposal set of \scione{4} samples. For the discrete variant, we perform 50 annealing steps, requiring two energy evaluations per step: one to update the samples and one for the Metropolis–Hastings correction. For the continuous variants, we perform 100 annealing steps. In both cases, resampling is performed at every step. For the continuous variant, Langevin dynamics is used with a step size \(\sigma_t\) of \scione{-7} for dipeptides and tetrapeptides, and  \(\sigma_t\) of \scione{-8} for octapeptides. Details of the formulation are found in \cref{app:cont-time-ais}. 
For the discrete variant, we apply Langevin dynamics with a step size of \scione{-5}, followed by a Metropolis-Hastings step to accept or reject proposals. The step size is adaptively updated to maintain an acceptance rate of approximately 60\%, under the assumption of sufficient smoothness in the intermediate densities. Further details of discrete-time AIS are found in \cref{app:discrete-time-ais}.

\paragraph{Self-improvement} We perform 4 rounds of self-improvement. In each round, we spend a portion of the budget to generate \sci{2}{5} samples and reweight using SNIS. The resulting reweighted samples are then used to finetune the model for 250 gradient steps with a batch size of 256. Notably, once the buffer is established, the finetuning does not further expend the allocated budget as it does not involve energy evaluations. After the final round, the remaining computational budget is allocated to generate a final set of \sci{2}{5} samples, which are again SNIS reweighted to yield the empirical distribution \(\tilde{p}(x | s)\) for a given system \(s\). We found it beneficial to introduce an $L_2$ regularization term between the log-densities of the current model and a “teacher” model initialized from the pre-trained weights.

\begin{figure}[h]
    \centering
    \includegraphics[width=0.5\linewidth]{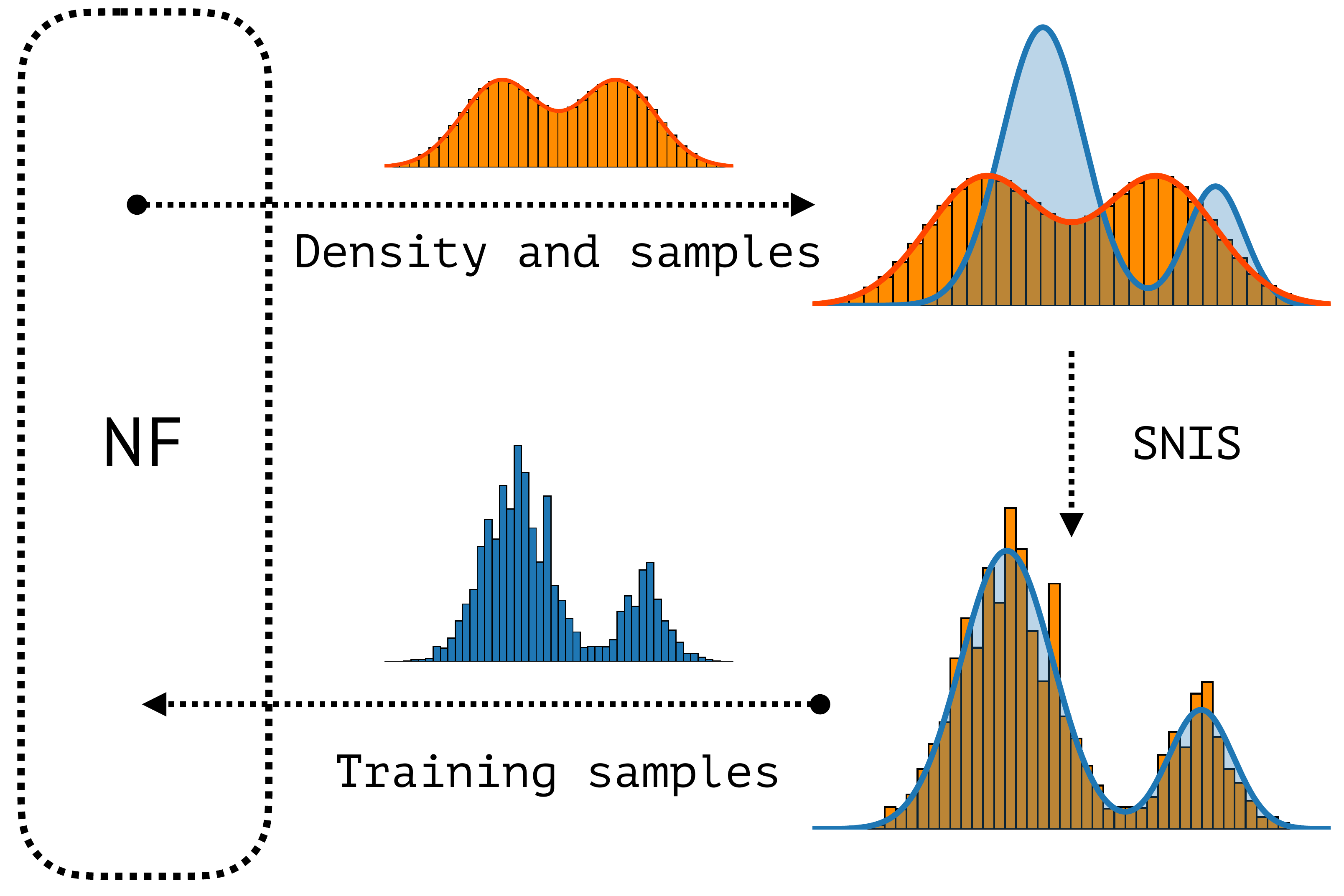}
    \caption{\textbf{Self-improvement procedure}. A pre-trained \name is finetuned at inference-time by iteratively generating samples, reweighting them using SNIS, and training on the reweighted samples.}
    \label{fig:self-refinement}
\end{figure}

\subsection{Additional baseline configurations}\label{app:baseline_configs}

\paragraph{TimeWarp} The original codebase and model weights of \citet{klein2023timewarp} were sampled using the asymptotically unbiased MCMC variant (with Metropolis-Hastings acceptance).

\paragraph{BioEmu} The inference code of \citet{lewis2024scalable} was employed. BioEmu does not directly model all-atom resolution hence \texttt{hpacker} is employed to introduce the side chains before energy minimization. The codebase of \citep{lewis2024scalable} was adapted to use the same Amber14 forcefield as ManyPeptidesMD, on which the \(10^4\) energy evaluation budget was spent on per-sample minimization. We additionally experimented with equilibration to target the larger \(10^6\) energy budget but could not achieve superior results to minimization-only hence this was not included. The adapted codebase for this baseline is provided at \url{https://github.com/transferable-samplers/BioEmu}.

\paragraph{UniSim} The UniSim model trained on the PepMD dataset of \citet{yu2025unisimunifiedsimulatortimecoarsened} was evaluated. UniSim applies energy minimization following a proposal step; simulation was ran until the \(10^4\) energy budget was expended. Increasing the simulation time to include the larger \(10^6\) energy evaluation budget was not found to improve performance and was omitted. The adapted codebase for this baseline is provided at \url{https://github.com/transferable-samplers/unisim}.

\subsection{Computational resources}

\begin{figure}[h]
    \centering
    \includegraphics[width=0.5\textwidth]{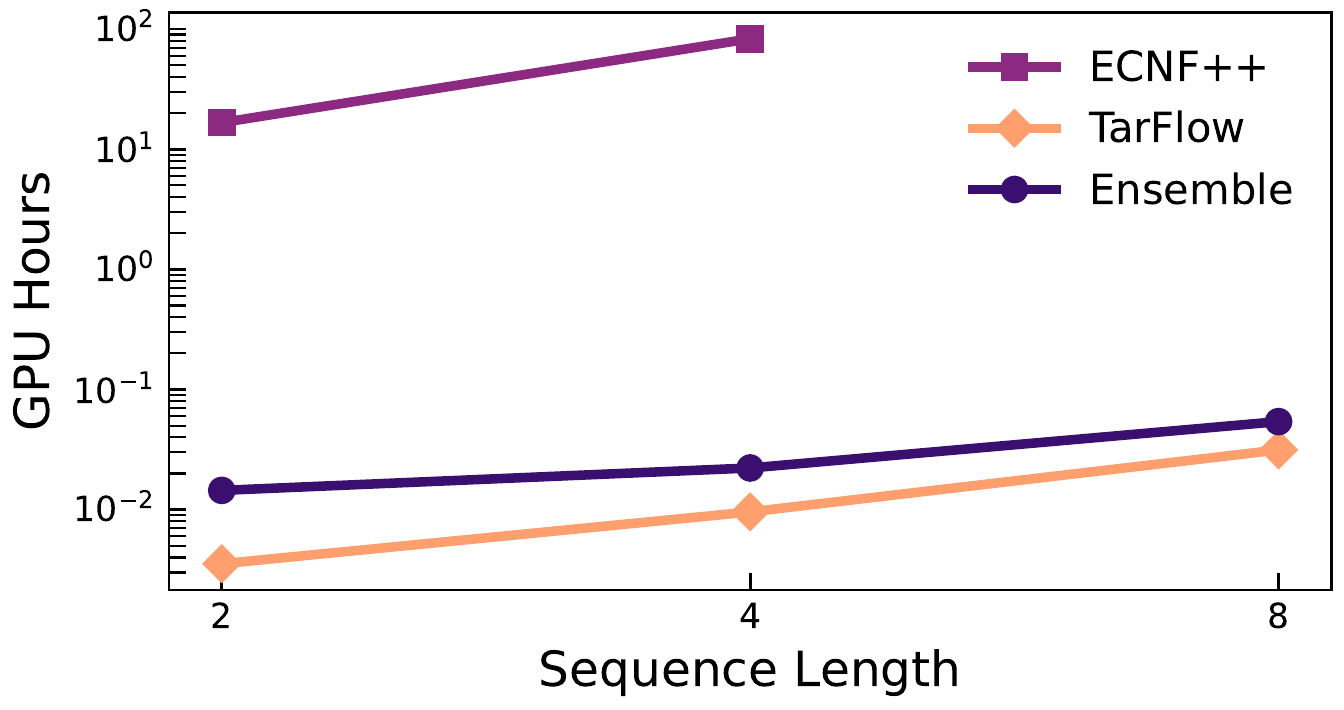}
    \caption{Sampling time for \scione{4} samples on NVIDIA L40S GPU for Boltzmann generators presented in \cref{tab:main_results}.}
    \label{fig:timings}
\end{figure}

All evaluation experiments are run on a heterogeneous cluster of NVIDIA L40S and RTX8000 GPUs. ECNF++ sampling is parallelized across multiple nodes with unique seeds to reduce sequential runtime. All evaluation timings are recorded using NVIDIA L40S GPUs. The sampling time required for \scione{4} samples for each model is presented in \cref{fig:timings}.

\section{Supplementary experimental results}
\label{app:additional_results}

\subsection{Unweighted proposal performance}
We compare the performance of ECNF++, TarFlow, and \name before and after SNIS (with \scione{4} samples) in \cref{tab:proposal_results}. Evidently, the strongest proposal distribution is given by ECNF++, with both TarFlow and \name having large values of \emetric due to high-energy samples. Notably, ECNF++ deteriorates on all metrics after reweighting, suggesting error accumulation in the divergence integration. TarFlow and \name improve significantly on \emetric after reweighting; however, only \name achieves a reduction in macrostructure metrics through reweighting. Whilst the unweighted ECNF++ proposal achieves stronger \torusmetric and \ticametric on dipeptides and tetrapeptides than SNIS-reweighted \name, this must be considered with the higher \emetric indicating poor local detail.

\begin{table*}[ht!]
\caption{\small Quantitative results for flows comparing the proposal performance and performance after importance sampling on peptide systems up to 8 residues. SNIS performed with a budget of \(10^4\) energy evaluations.}
\label{tab:proposal_results}
\resizebox{1\linewidth}{!}{
\begin{tabular}{@{}llccccccccc@{}}
\toprule
Sequence length $\rightarrow$ & & \multicolumn{2}{c}{2AA \tiny{(30 systems)}} & \multicolumn{3}{c}{4AA \tiny{(30 systems)}} &
\multicolumn{3}{c}{8AA \tiny{(30 systems)}}\\
\cmidrule(lr){3-4} \cmidrule(lr){5-7}
\cmidrule(lr){8-10}
Model $\downarrow$ & & \emetric $\downarrow$ & \torusmetric $\downarrow$ & \emetric $\downarrow$ & \torusmetric $\downarrow$ & \ticametric $\downarrow$ & \emetric $\downarrow$ & \torusmetric $\downarrow$ & \ticametric $\downarrow$\\
\midrule
\multirow{2}{*}{ECNF++} 
  & Proposal & 1.958 & 0.143 & 5.006 & 0.582 & 0.335 & --- & --- & --- \\
  & SNIS & 3.470 & 0.302 & 10.032 & 1.121 & 0.572 & --- & --- & --- \\
\midrule

\multirow{2}{*}{TarFlow} 
  & Proposal &  \(>10^6\) & {0.178} & \(>10^9\) & {0.882} & {0.384} & \(>10^9\) & 2.475 & 1.026\\
  & SNIS & 0.452 & {0.193} & 1.260 & 0.924 & 0.492 & 11.298 & 2.733 & 1.087 \\
\midrule

\multirow{2}{*}{\name} 
  & Proposal & \(>10^9\) & 0.261 & \(>10^9\) & 0.916 & 0.546 & \(>10^9\) & 2.456 & 1.081\\
  & SNIS     & 0.371 & 0.210 & 0.932 & {0.752} & {0.367} & 10.038 & {2.456} & 0.988 \\
\bottomrule
\end{tabular}
}
\vspace{-5pt}
\end{table*}

\subsection{Additional ablations}

We consider three further ablations of the \name architecture. Firstly, we replace the adaptive conditioning and transition blocks with (i) deeper transformation blocks in which the 8 transformer layers are increased to 20 (ii) a wider transformer block in which the dimension is increased from 384 to 576. In both cases the increased depth / width was defined to approximately match the parameter count of \name. We additionally ablate the use of \emph{lookahead conditioning}, in which the atom token $z[i]$ is conditioned not only on $[A_i,R_i,P_i,L_i]$ but also $[A_{i+1},R_{i+1},P_{i+1},L_{i+1}]$, with conditioning information for this pair of indexes mixed using a small MLP. This was motivated by the observation that given naive conditioning the causal masking implies updates to $z[i]$ are computed without knowledge of $[A_i,R_i,P_i,L_i]$. Preliminary experiments up to tetrapeptides suggested look-ahead conditioning to be beneficial hence it was included in the final \name model.

In \cref{tab:extra_ablation_results} we present results for these ablation models, trained using the same procedure as \name. We observe \name to marginally outperform both w/o transition variants, while removing the look-ahead conditioning is in fact beneficial, particularly on the shorter sequences. These results invite further research into the optimal allocation of parameters, and advanced conditioning techniques for transferable autoregressive flows.

\begin{table*}[ht!]
\caption{\small SNIS is performed with a fixed budget of \(2 \times 10^5\) energy evaluations.}
\label{tab:extra_ablation_results}
\resizebox{1\linewidth}{!}{
\begin{tabular}{@{}lccccccccccc}
    \toprule
    Sequence length $\rightarrow$ & \multicolumn3c{2AA \tiny{(30 systems)}} & \multicolumn4c{4AA \tiny{(30 systems)}} & \multicolumn4c{8AA \tiny{(30 systems)}}  \\
    \cmidrule(lr){2-4}\cmidrule(lr){5-8}\cmidrule(lr){9-12}
    Model $\downarrow$ & ESS $\uparrow$ & \emetric  $\downarrow$  & \torusmetric  $\downarrow$  & ESS $\uparrow$ & \emetric  $\downarrow$  & \torusmetric  $\downarrow$  & \ticametric  $\downarrow$  & ESS $\uparrow$ & \emetric  $\downarrow$  & \torusmetric  $\downarrow$  & \ticametric  $\downarrow$  \\
    \midrule
    \name & {0.191} & {0.282} & 0.177 & {0.071} & 0.646 & {0.607} & 0.349 & {0.011} & 9.360 & {2.019} & {0.960} \\
    \midrule
    w/o transition (deep) & 0.166 & 0.290 & {0.154} & 0.060 & 0.643 &{0.613} & {0.338} & 0.010 & {9.257} & 2.123 & {0.961} \\
    w/o transition (wide) & 0.187 & 0.291 & 0.158 & 0.065 & {0.637} & 0.634 & 0.354 & 0.010 & 9.426 & 2.121 & 0.988 \\
    w/o lookahead & {0.212} & {0.270} & {0.158} & {0.074} & {0.591} & {0.623} & {0.372} & {0.011} & {9.319} & {2.070} & {0.962} \\
    \bottomrule
    \end{tabular}
    }
    \vspace{-5pt}
\end{table*}

\subsection{Dataset scaling}
\label{app:datascale}

We explore the effect of dataset size on the performance of \name. The full ManyPeptidesMD dataset contains 21,700 sequences simulated for \SI{200}{\ns} each. We train a further four models wherein (i) the trajectories are limited to \SI{50}{\ns} and \SI{100}{\ns} respectively, and (ii) the set of sequences is reduced to 25\% and 50\%. For the 25\% and 50\% sequence reduction variants sequence reduction is applied non-linearly with a simple estimate of relative cost, to avoid excessively removing the relatively cheap simulations of short sequences. In \cref{fig:scaling-plot} we present metrics for these models. We note the \emetric to lack any meaningful trend, and the \torusmetric to have limited sensitivity to the dataset variation, despite a slight trend towards larger data subsets improving the metrics. Most interesting of these plots is the \ticametric with a clear trend indicating the full \SI{200}{ns} trajectories to be beneficial, whilst using only 50\% of the sequences is in fact superior to the full dataset. We caveat these results with the observation that \name may indeed be operating in the compute-bound regime; even subsampled to \SI{10}{\ps} per frame the  ManyPeptidesMD training dataset contains sufficient data for over \sci{8}{5} training iterations (batch size 512) without repeating a single data sample, in excess of the training budget of \sci{5}{5} permitted to the \name models in this paper.

\begin{figure}[ht!]
    \centering
    \includegraphics[width=0.95\linewidth]{media/data_scaling.pdf}
    \caption{Wasserstein-distance metrics for \name trained on variants of ManyPeptidesMD. Upper row: ManyPeptidesMD contains \SI{200}{\ns} trajectories for all training sequences, we train on a variant with \SI{50}{\ns} and \SI{100}{\ns} respectively. Lower: ManyPeptidesMD contains 21,700 sequences, we train using 25\% and 50\% of the total. Evaluation metrics computed using SNIS on 30 octapeptide sequences.}
    \label{fig:scaling-plot}
\end{figure}

\subsection{Sequence-length extrapolation}

Recall that \name is trained exclusively on peptide sequences up to length eight. We explore the capacity of the model to generalize in sequence length beyond its training distribution by evaluating on the nine-residue sequence \texttt{YQNPDGSQA} described by \citet{wu_beta-hairpin_2004} and the well-studied ten-residue small protein Chignolin \texttt{GYDPETGTWG} \citep{honda200410}. We additionally evaluate BioEmu and UniSim as baselines on these systems. 

\begin{table}[ht!]
\centering
\caption{Comparison of metrics for \texttt{YQNPDGSQA} (9AA) and Chignolin / \texttt{GYDPETGTWG} (10AA).}
\resizebox{1\linewidth}{!}{
\begin{tabular}{@{}lccccccccc}
\toprule
& & \multicolumn{4}{c}{\texttt{YQNPDGSQA} (9AA)} & \multicolumn{4}{c}{Chignolin (10AA)} \\
\cmidrule(lr){3-6} \cmidrule(lr){7-10}
 & & ESS & \emetric & \torusmetric & \ticametric & ESS & \emetric & \torusmetric & \ticametric \\
\midrule
UniSim &  & -- & $>10^5$ & 6.00 & 0.84 & -- & 267.68 & 6.48 & 0.20 \\
BioEmu & & -- & 160.52 & 4.52 & 1.14 & -- & 198.90 & 5.14 & 0.65 \\
\midrule
\name & Proposal & -- & \(>10^9\) & 3.91 & 1.65 & -- & \(>10^9\) & 3.63 & 0.96 \\
\name & SNIS & 0.0049 & 23.79 & 3.85 & 1.94 & 0.0001 & 832.59 & 4.35 & 1.25 \\
\name (self-improve) & Proposal & -- & \(>10^9\) & 3.73 & 1.95 & -- & \(>10^9\) & 3.94 & 1.13 \\
\name (self-improve) & SNIS & 0.0123 & 18.85 & 3.79 & 1.95 & 0.0002 & 275.87 & 4.43 & 1.20 \\
\bottomrule
\end{tabular}
}
\label{tab:ensemble_results}
\end{table}

\subsection{JSD evaluation}
To further assess distributional alignment between generated samples and samples from the reference MD, we follow \citet{raja2025actionminimization} and compute the Jensen–Shannon divergence (JSD) across both TICA projections and backbone torsion angles; for more details on this metric see \cref{app:metrics}

\begin{table*}[ht!]
\caption{\small Quantitative results for flows comparing the \(\mathrm{JSD}\) performance on \(\mathrm{TICA}\) projections and torus angles before and after importance sampling. SNIS performed with a budget of \(10^4\) energy evaluations.}
\label{tab:jsd_results}
\resizebox{1\linewidth}{!}{
\begin{tabular}{@{}llccccc@{}}
\toprule
Sequence length $\rightarrow$ & & \multicolumn{1}{c}{2AA \tiny{(30 systems)}} & \multicolumn{2}{c}{4AA \tiny{(30 systems)}} &
\multicolumn{2}{c}{8AA \tiny{(30 systems)}}\\
\cmidrule(lr){3-3} \cmidrule(lr){4-5} \cmidrule(lr){6-7}
Model $\downarrow$ & & \torusjsd $\downarrow$ & \torusjsd $\downarrow$ & \ticajsd $\downarrow$ & \torusjsd $\downarrow$ & \ticajsd $\downarrow$\\
\midrule
    TimeWarp & --- & 0.280 & 0.460 & 0.415 & --- & --- \\
    BioEmu & --- & 0.329 & 0.245 & 0.315 & 0.371 & 0.403 \\
    UniSim & --- & 0.381 & 0.586 & 0.376 & 0.879 & 0.609 \\
\midrule
\multirow{2}{*}{ECNF} 
  & Proposal & 0.007 & --- &  --- & --- & --- \\
  & SNIS & 0.031 & ---- & --- & --- & --- \\
\midrule
\multirow{2}{*}{ECNF++} 
  & Proposal & {0.002} & {0.004} & 0.004 & --- & --- \\
  & SNIS & 0.020 & 0.051 & 0.052 & --- & --- \\
\midrule
\multirow{2}{*}{TarFlow} 
  & Proposal & 0.006 & 0.034 & 0.017 & 0.104 & 0.098  \\
  & SNIS & 0.005 & 0.022 & 0.019 & 0.139 & 0.124 \\
\midrule
\multirow{2}{*}{\name} 
  & Proposal & 0.006 & 0.027 & 0.023 & 0.095 & {0.091}  \\
  & SNIS & 0.004 & 0.011 &{0.009} & 0.109 & 0.082 \\
\bottomrule
\end{tabular}
}
\vspace{-5pt}
\end{table*}

\subsection{Effective sample per second}

We report the effective sample size per second (ESS/s) for \name and baseline Boltzmann generators, evaluated using SNIS with \scione{4} energy evaluations on an NVIDIA L40s GPU.

\begin{table}[h]
\centering
\caption{Effective sample size per second (ESS/s)}
\begin{tabular}{lccc}
\toprule
 & 2AA & 4AA & 8AA \\
\midrule
ECNF & $1.59\cdot10^{-2}$ & --- & --- \\
ECNF++ & $4.76\cdot10^{-3}$ & $2.78\cdot10^{-4}$ & --- \\
TarFlow & $1.48\cdot10^{2}$ & $2.50\cdot10^{1}$ & $1.31\cdot10^{1}$ \\
\name & $7.58\cdot10^{1}$ & $1.64\cdot10^{1}$ & $9.86\cdot10^{-1}$ \\
\bottomrule
\end{tabular}
\end{table}

\clearpage
\subsection{Octapeptide Ramachandran plots}
\begin{figure}[h!]
\centering
\includegraphics[width=0.75\textwidth]{media/octopeptide_rama.pdf}
\caption{Ramachandran plots for \texttt{DGVAHALS} unseen octapeptide system. Reference molecular dynamics (left column), \name proposal (center column), \name SNIS with \scione{5} samples (right column).}
\end{figure}

\clearpage
\subsection{Temperature plots}
\begin{figure}[h]
    \centering
    \includegraphics[width=\linewidth]{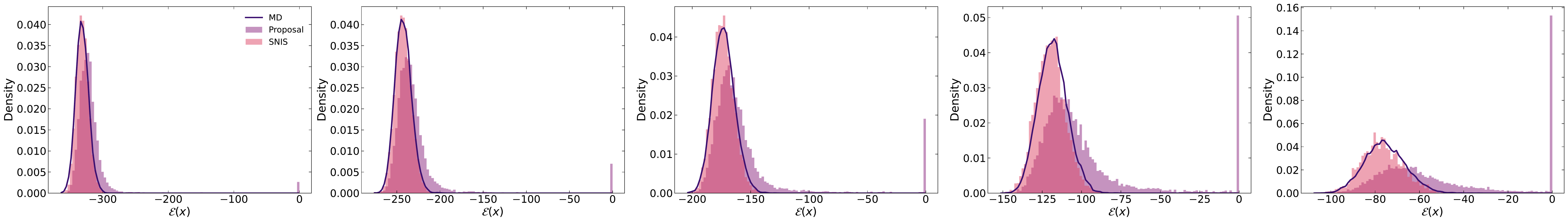}
    \caption{Energy histogram plots for \tempseq at different temperatures.}
    \label{fig:energy-temp}
\end{figure}

We present TICA plots in \cref{fig:tica-temp} and energy distributions in \cref{fig:energy-temp} across a range of temperatures (310K, 393K, 498K, 631K, 800K). At each temperature, we generate \sci{2}{5} samples by scaling the prior with the inverse temperature \(\beta\), sampling from \(\mathcal{N}(0, 1/\beta)\). For SNIS, we use the energy at the corresponding temperature to reweight the samples.

\begin{figure}[h]
    \centering
    \includegraphics[width=\linewidth]{media/RLMM_temp_ticas.pdf}
    \caption{TICA plots for \tempseq at different temperatures. Reference molecular dynamics (top row), \name proposal (middle row), \name SNIS (bottom row) with \sci{5}{4} samples.}
    \label{fig:tica-temp}
\end{figure}

\end{document}